\documentclass[journal]{IEEEtai}
\usepackage[colorlinks,urlcolor=blue,linkcolor=blue,citecolor=blue]{hyperref}
\usepackage{times}
\usepackage{epsfig}
\usepackage{graphicx}
\usepackage{amsmath}
\usepackage{amssymb}
\usepackage{url}
\usepackage{algorithmic}
\usepackage{algorithm}

\usepackage{booktabs} 
\usepackage{enumerate}
\usepackage{adjustbox}
\usepackage{indentfirst}
\usepackage{array}
\usepackage{multirow}
\usepackage{float}
\usepackage{hhline}
\usepackage{bm}
\usepackage{subcaption}
\usepackage{color}

\usepackage{fancyhdr}
\fancypagestyle{firstpage}{%
\lhead{IEEE Transactions on Artificial Intelligence}
\rhead{}
  \lfoot{\small{\copyright2020 IEEE. 
This paper is accepted in IEEE Transactions on Artificial Intelligence. Copyright has been transferred to IEEE. Personal use of this material is permitted. Permission from IEEE must be obtained for all other users, including reprinting/republishing this material for advertising or promotional purposes, creating new collective works for resale or redistribution to servers or lists, or reuse of any copyrighted components of this work in other works.}}
}

\begin{document}

\title{Color Channel Perturbation Attacks for Fooling Convolutional Neural Networks and A Defense Against Such Attacks}

\author{Jayendra Kantipudi, Shiv Ram Dubey, \IEEEmembership{Member,~IEEE}, and Soumendu Chakraborty
\thanks{K. Jayendra and S.R. Dubey are with Computer Vision Group at Indian Institute of Information Technology, Sri City, Chittoor, Andhra Pradesh, India (Email: jayendra.k17@iiits.in, srdubey@iiits.in).}
\thanks{S. Chakraborty is with Indian Institute of Information Technology, Lucknow, Uttar Pradesh, India (Email: soumendu@iiitl.ac.in).}
}

\markboth{IEEE Transactions on Artificial Intelligence}
{Jayendra Kantipudi \MakeLowercase{\textit{et al.}}}

\maketitle
\thispagestyle{firstpage}

\begin{abstract}
The  Convolutional  Neural  Networks  (CNNs) have emerged as a very powerful data dependent hierarchical feature extraction method. It is widely used in several computer vision problems. The CNNs learn the important visual features from training samples automatically. It is observed that the network overfits the training samples very easily. Several regularization methods have been proposed to avoid the overfitting. In spite of this, the network is sensitive to the color distribution within the images which is ignored by the existing approaches. In this paper, we discover the color robustness problem of CNN by proposing a Color Channel Perturbation (CCP) attack to fool the CNNs. In CCP attack new images are generated with new channels created by combining the original channels with the stochastic weights. Experiments were carried out over widely used CIFAR10, Caltech256 and TinyImageNet datasets in the image classification framework. The VGG, ResNet and DenseNet models are used to test the impact of the proposed attack. It is observed that the performance of the CNNs degrades drastically under the proposed CCP attack. Result show the effect of the proposed simple CCP attack over the robustness of the CNN trained model. The results are also compared with existing CNN fooling approaches to evaluate the accuracy drop. We also propose a primary defense mechanism to this problem by augmenting the training dataset with the proposed CCP attack. The state-of-the-art performance using the proposed solution in terms of the CNN robustness under CCP attack is observed in the experiments. The code is made publicly available at \url{https://github.com/jayendrakantipudi/Color-Channel-Perturbation-Attack}.
\end{abstract}

\begin{IEEEImpStatement}
Convolutional Neural Networks (CNNs) are an important feature learning mechanism for visual data.
They have been very successful in many computer vision applications, primarily within a controlled experimental setup. 
However, CNNs can be fooled very easily by augmenting the test data, which limits its practical uses over unseen data in real applications.
We propose a simple image color channel perturbation (CCP) attack over test data, which fools the CNNs and reduces the average accuracy by $41\%$. 
We also use the CCP based data augmentation during training to increase the robustness of the CNNs, which achieves an increase in average accuracy by $63\%$. 
The proposed approach can be used to improve the robustness of CNNs against the attacks over the color channels for different vision based intelligent applications.
\end{IEEEImpStatement}


\section{Introduction}
\begin{figure}[!t]
\centering
\includegraphics[trim=5 110 58 30, width=\columnwidth, clip]{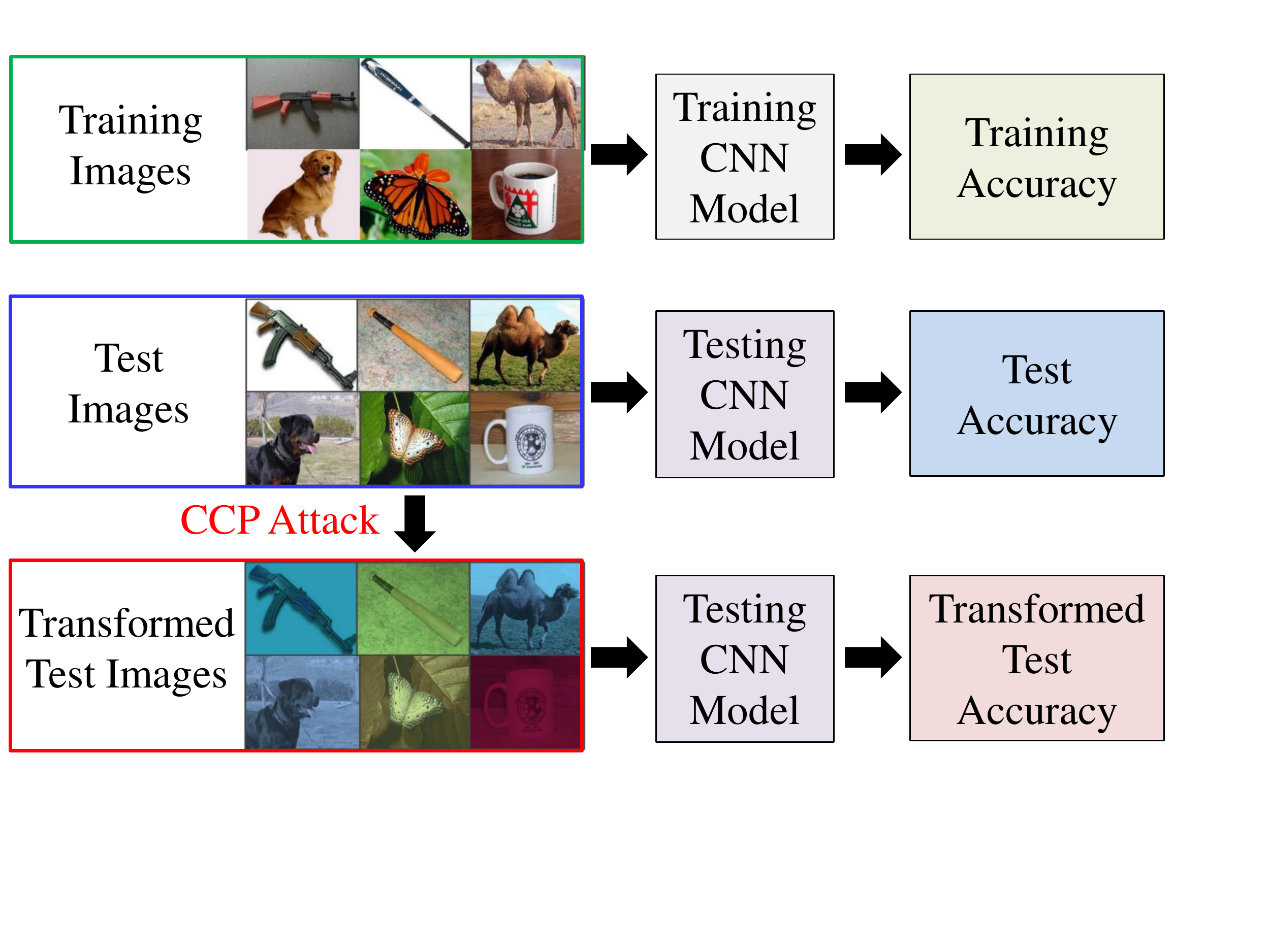}
\caption{The illustration of the proposed Color Channel Perturbation (CCP) attacks to fool the CNN for the object recognition. The images are taken from the Caltech256 dataset \cite{caltech256}. The $1^{st}$ row represents the training over training set, $2^{nd}$ row represents the testing over test set, and $3^{rd}$ row represents the testing over the transformed test set using the proposed CCP attack.}
\label{fig:idea}
\end{figure}

\IEEEPARstart{T}{he} last decade was completely dominated by the deep learning methods to solve the various problems of Computer Vision, Natural Language Processing, Robotics, and others \cite{liu2017survey}. Deep learning methods learn the important features from the data automatically in a hierarchical fashion \cite{lecun2015deep}. It has shown very promising performance in various domains, such as computer vision \cite{ioannidou2017deep}, \cite{guo2016deep}, natural language processing \cite{young2018recent}, \cite{strubell2019energy}, health informatics \cite{ravi2016deep}, \cite{shen2017deep}, sentiment analysis from social media data \cite{glorot2011domain}, \cite{severyn2015twitter}, etc. However, the recent studies show that the trained deep learning models can be fooled easily by manipulating the test data \cite{goodfellow2014explaining}, \cite{moosavi2016deepfool}, \cite{elsayed2018adversarial}, \cite{su2019one}, \cite{zhao2019adversarial}. Most of the existing fooling techniques are data dependent, whereas we present a data independent color channel perturbation based attack in this paper to fool the trained deep learning models as depicted in Fig. \ref{fig:idea}.

The Convolutional Neural Network (CNN) is a type of neural network designed to incorporate deep learning in the classification of image and video data. The AlexNet was the first popular CNN model developed by Krizhevsky et al. in 2012 \cite{alexnet}. It won the ImageNet object recognition challenge in 2012 and shown a great improvement \cite{imagenet}.
Inspired from the success of AlexNet, various CNN models have been investigated for object recognition, including VGG \cite{vgg}, ResNet \cite{resnet}, DenseNet \cite{densenet}, etc.
The CNN based models have been also proposed for different applications. The Faster R-CNN \cite{fasterrcnn}, Single Shot Detector (SSD) \cite{ssd} and You Only Look Once (YOLO) \cite{yolo} are proposed for object detection. The Mask R-CNN \cite{he2017mask} is designed for segmentation and the MicroExpSTCNN \cite{reddy2019spontaneous} is proposed for micro-expression recognition. The HybridSN \cite{roy2019hybridsn} is proposed for hyperspectral image classification. There are CNN applications, which include image classification \cite{basha2020impact}, face recognition \cite{srivastava2019performance}, \cite{srivastava2019hard}, face anti-spoofing \cite{nagpal2019performance}, image-to-image transformation \cite{kancharagunta2019csgan}, \cite{babu2020cdgan}, and many more.

The basic aim of any CNN is to learn the important features automatically from the training data. The learning of the weights of CNN is generally done using the back propagation technique \cite{adam}, \cite{dubey2019diffgrad}. Normally, the CNNs are likely to overfit the training data if the complexity of the CNN model is higher than the dataset and proper regularizations are not used. Some common regularization techniques are Dropout \cite{dropout}, Batch Normalization \cite{batchnormalization}, Data Augmentation \cite{dataaugmentation}, etc.

\begin{figure*}[!t]
\centering
\includegraphics[trim=5 30 5 10, width=\linewidth, clip]{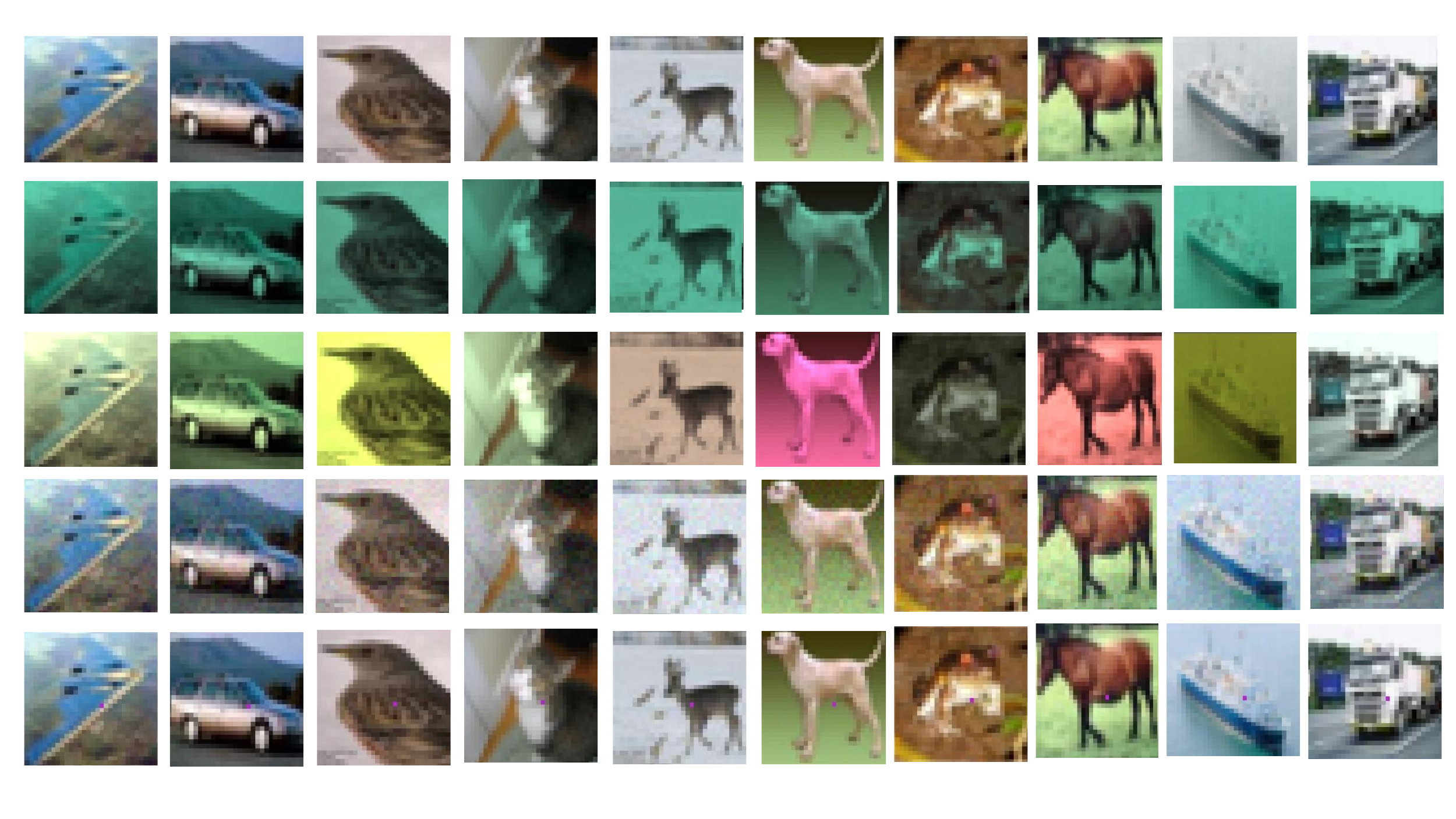}
\caption{Sample images from ($1^{st}$ row) original test set of CIFAR10 dataset \cite{cifar}, ($2^{nd}$ row) transformed using CCP attack with fixed random weight setting, ($3^{rd}$ row) transformed using CCP attack with variable random weight setting, ($4^{th}$ row) transformed using Adversarial attack \cite{goodfellow2014explaining}, and ($5^{th}$ row) transformed using One-pixel attack \cite{su2019one}. }
\label{fig:sample}
\end{figure*}

It has been observed in the literature that the trained CNN models can be fooled by modifying the test data using different approaches \cite{goodfellow2014explaining}. Recent studies show that the CNNs are vulnerable to maliciously designed perturbations (i.e., adversarial examples) \cite{moosavi2016deepfool}, \cite{elsayed2018adversarial}, \cite{moosavi2017universal}. 
Su et al. have generated the differential
evolution based One-pixel adversarial perturbations to fool the CNNs \cite{su2019one}.
Zhao et al. have introduced a one-step spectral attack (OSSA) using the Fisher information in neural networks \cite{zhao2019adversarial}. 
The adversarial transformation using the shape bias property is developed to generate the semantic adversarial examples to fool the CNNs \cite{hosseini2018semantic}. Joshi et al. have also developed the semantic adversarial attacks using the parametric transformations to fool the CNNs \cite{joshi2019semantic}. Bhattad et al. have developed the texture transfer and colorization based big, but imperceptible adversarial perturbations \cite{bhattad2019big}. A functional adversarial attack is also investigated to fool the machine learning models \cite{laidlaw2019functional}. Guo et al. have generated the adversarial images in the low frequency domain \cite{guo2018low}. It is observed in the literature that the adversarial attack is widely explored to fool the CNNs. However, most of the existing methods do not attack the color channels, which is exploited by the proposed color channel perturbation attack.

In order to cope up with the problem of CNNs getting fooled by the synthesized test images, several defense mechanisms and robustness properties of the network have been explored.
Billovits et al. have observed that the robust CNNs can be developed for adversarial attack by preserving the L2 norm of the original image in the corresponding adversarial image \cite{billovits2016hitting}.
Feng et al. have analyzed the effect of adversarial attacks over a Deep product quantization network (DPQN) for image retrieval \cite{feng2020adversarial}.
Agarwal et al. have used Support Vector Machine as the classifier coupled with the Principal Component Analysis as features for the detection of image-agnostic universal perturbations \cite{agarwal2018image}. 
Zheng et al. have utilized the CNN's intrinsic properties to detect adversarial inputs \cite{zheng2018robust}.
Prakash et al. have corrupted the image by redistributing pixel values using pixel deflection to increase the CNN robustness \cite{prakash2018deflecting}.
Raff et al. have shown that, combining a large number of individually weak defenses stochastically yields a strong defense against adversarial attacks \cite{raff2019barrage}.
Mao et al. have used the metric learning to produce robust classifiers against adversarial attack \cite{mao2019metric}.
We use data augmentation where new images are created by CCP attack to increase the robustness of the CNN models against CCP attack.

The existing methods focus over the adversarial attacks which use the gradient information and defense mechanism. However, to the best our knowledge the attack due to the color channel perturbation is not studied so far. 
Thus, in this paper, we introduce a stochastic color channel perturbation (CCP) attack to fool the CNNs as shown in Fig. \ref{fig:idea}. We also present an analysis of the proposed CCP attack when used in data augmentation during the training of different CNN models over different datasets.
Most of the existing image attack strategies have the following two major limitations: (a) they are data dependent and make use of network for attack which limits its applications in unknown scenarios, whereas the proposed attack strategy is independent of data, and (b) they are unable to create variations in the color, whereas the main aim of the proposed approach is to introduce the color perturbations while preserving the semantic meaning.
Following are the main contributions of this paper:
\begin{itemize}
  \item We propose a new data independent stochastic attack termed as Color Channel Perturbation (CCP) attack.
  \item The proposed CCP attack generates the color channels by combining the original color channels with stochastic weights.
  \item Unlike other attacks, the proposed CCP attack makes use of the same weights for each pixel of the image, i.e., uniform relative transformation within an image.
  \item We show the performance degradation in VGG, ResNet and DenseNet models due to the proposed attack over different benchmark datasets, including CIFAR10, Caltech256 and TinyImageNet.
  \item We analyze the performance of the proposed attack over both low and high resolution images.
  \item We also train the models with the synthesized images to observe the defense capability when the proposed attack is used as the data augmentation.
\end{itemize}

The rest of the paper is structured as follows: Section II presents the proposed color channel perturbation attack; Section III is devoted to experimental settings in terms of the CNNs used, datasets used and training settings followed; Section IV illustrates the experimental results and analysis, and finally the conclusions are drawn in Section V.

\begin{figure*}[!t]
\centering
\includegraphics[clip=true, trim = 14 95 22 40, width=\textwidth]{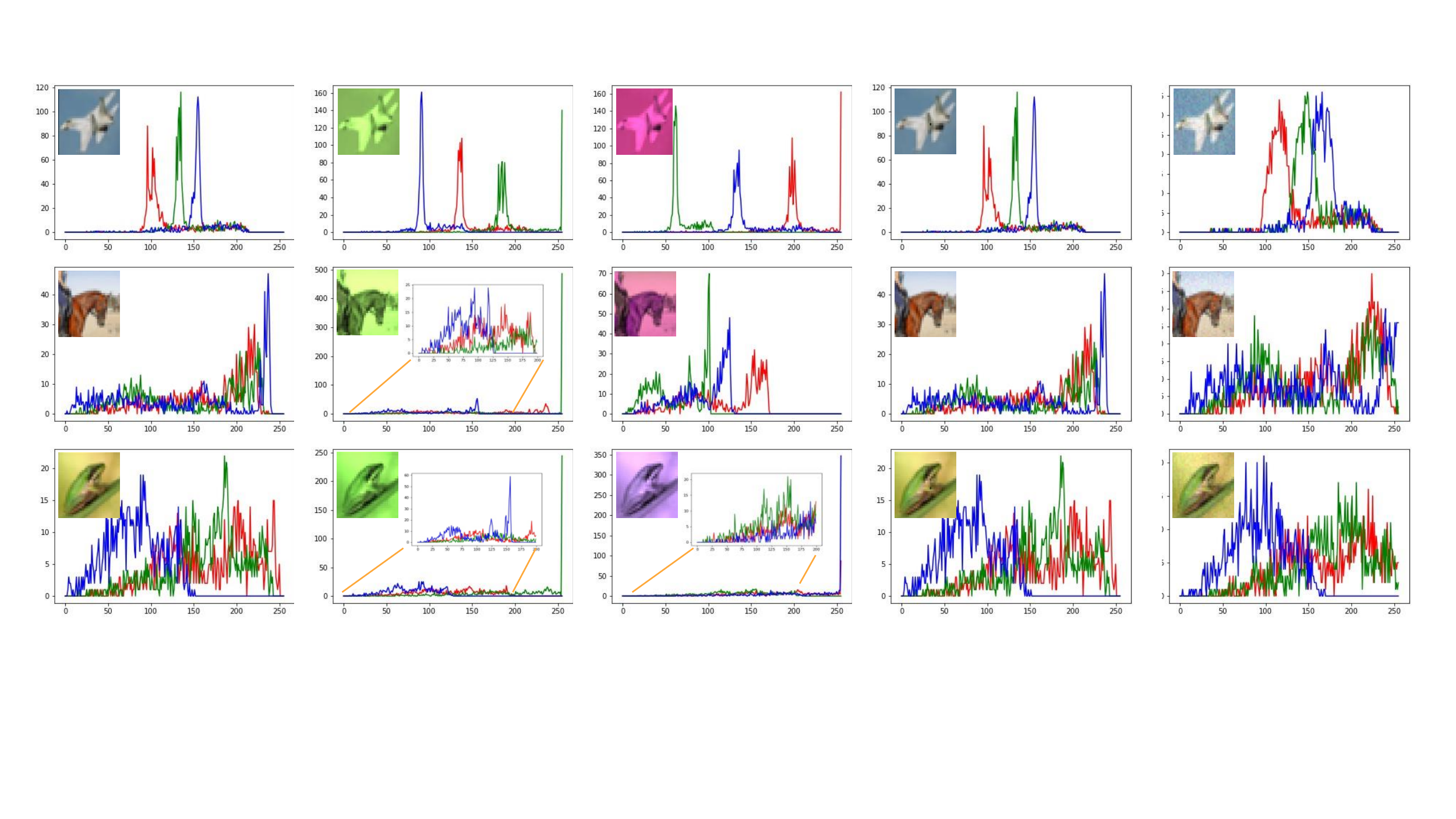}
\\
\includegraphics[clip=true, trim = 12 105 22 35, width=\textwidth]{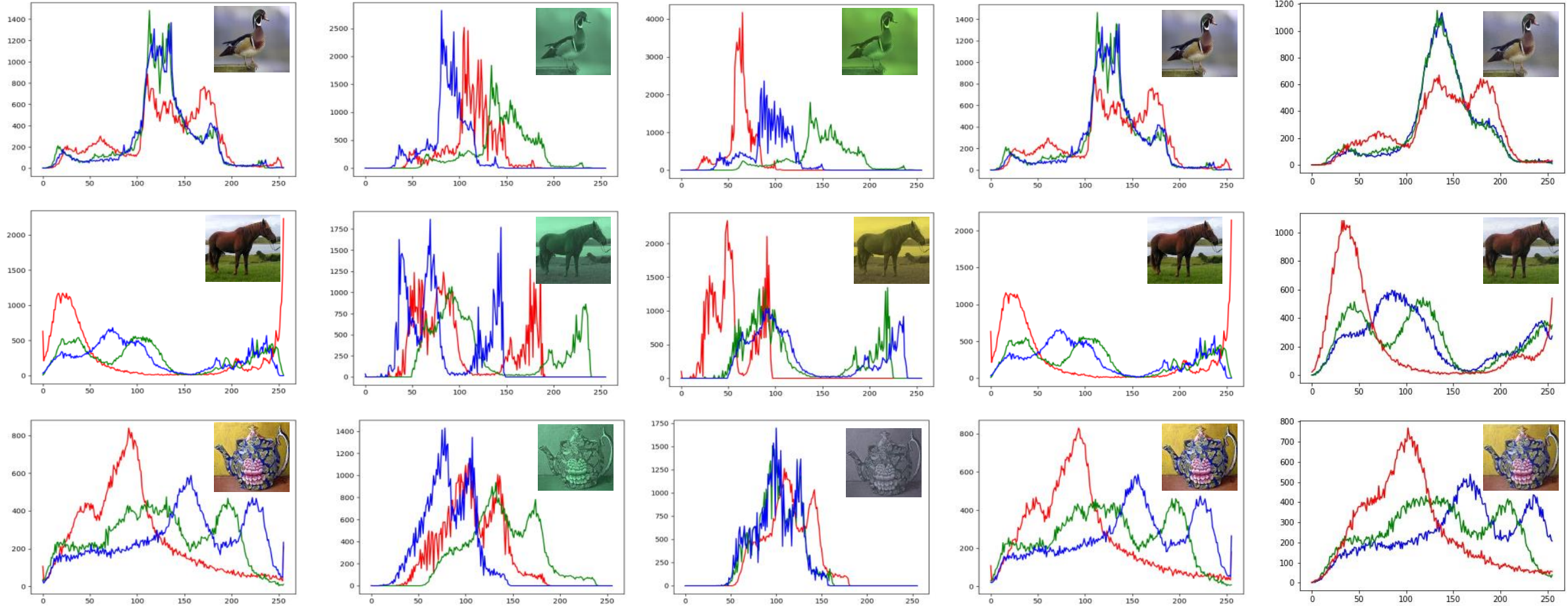}
\caption{The histograms of Red, Green and Blue channels computed under No attack, CCP attack with fixed random weight setting, CCP attack with variable random setting, One-pixel attack \cite{su2019one}, and Adversarial attack \cite{goodfellow2014explaining} over few images of the test set from ($1^{st}$ - $3^{rd}$ rows) CIFAR-10 dataset \cite{cifar} and ($4^{th}$ - $6^{th}$ rows) Caltech-256 dataset \cite{caltech256}.}
\label{fig:hist}
\end{figure*}

\section{Proposed Color Channel Perturbation Attack}
Recently, the researchers have observed that the trained Convolutional Neural Network (CNN) can be fooled by attacking the test image. However, the synthesized test image preserves the semantic meaning of the original test image. The performance of the CNN’s giving the correct result for the original test image degrades suddenly over the synthesized test image. Different types of attack methods have been explored in recent days. However, the existing attacking methods fail to utilize the color property of the image. 

In this paper, we propose a simple, yet effective Color Channel Perturbation (CCP) attacks on the image data to fool the CNNs. The proposed attack is based on the color property of the image. Any color image ($I$) contains three channels, namely Red ($R$), Green ($G$), and Blue ($B$) in RGB color space. The proposed CCP attack uses the original color channels (i.e., Red ($R$), Green ($G$), and Blue ($B$)) of the image ($I$) to generate the new transformed color channels (i.e., transformed Red ($R^T$), transformed Green ($G^T$), and transformed Blue ($B^T$)) of the transformed image ($I^T$). Basically, each one of $R^T$, $G^T$, and $B^T$ is the weighted combination of $R$, $G$, and $B$ and given as,
\begin{equation}
    R^T_i = s \times \left( \frac{\alpha^r_i \times R_i + \alpha^g_i \times G_i + \alpha^b_i \times B_i}{3} \right) + b
    \label{eqr}
\end{equation}
\begin{equation}
    G^T_i = s \times \left( \frac{\beta^r_i \times R_i + \beta^g_i \times G_i + \beta^b_i \times B_i}{3} \right) + b
    \label{eqg}
\end{equation}
\begin{equation}
    B^T_i = s \times \left( \frac{\gamma^r_i \times R_i + \gamma^g_i \times G_i + \gamma^b_i \times B_i}{3} \right) + b
    \label{eqb}
\end{equation}
where $s$ is a scale factor hyperparameter, $b$ is a bias hyperparameter, $\{R_i, G_i, B_i\}\in\mathbb{R}$ are the intensity values in Red, Green and Blue channels of $i^{th}$ input image ($I_i$) of the test set (i.e., $i \in [1, N_{test}]$ where $N_{test}$ is the number of images in the test set), $\{R^T_i, G^T_i, B^T_i\}\in\mathbb{R}$ are the intensity values in Red, Green and Blue channels of the transformed image ($I^T_i$) corresponding to $I_i$, $\alpha_i=\{\alpha^r_i, \alpha^g_i, \alpha^b_i\}$, $\beta_i=\{\beta^r_i, \beta^g_i, \beta^b_i\}$ and $\gamma_i=\{\gamma^r_i, \gamma^g_i, \gamma^b_i\}$ are the random weights to generate the Red, Green and Blue channels of the transformed image, respectively, corresponding to $i^{th}$ input image with $\{\alpha^\eta_i, \beta^\eta_i, \gamma^\eta_i \} \in [L, U]$, $\forall{\eta \in \{r, g, b\}}$ where $L$ and $U$ are the lower and upper limits of random numbers/weights generated, respectively. Note that we use $L=0$ and $U=1$ in the experiments until or otherwise specified. The scale $s$ and bias $b$ are used to adjust the visual appearance of the generated image. 

We propose two schemes of the proposed CCP attack, namely fixed random weight based CCP attack ($CCP_F$) and variable random weight based CCP attack ($CCP_V$). Note that in both the cases, the random weights generated are same for all the pixels of an image to preserve the relative local neighborhood information in the image. However, the random weights for different channels are generated independently in both the cases. In fixed scheme $CCP_F$, the random weight generated is same for all the test images, i.e., $\alpha^\eta_i = \alpha^\eta_j$, $\beta^\eta_i = \beta^\eta_j$ and $\gamma^\eta_i = \gamma^\eta_j$, $\forall{\{i, j\} \in [1, N_{test}]}$ and $\forall{\eta \in \{r, g, b\}}$.
However, in the variable scheme $CCP_V$, independent random weights are generated for different test images, i.e., the weights $\alpha^\eta_i$, $\beta^\eta_i$ and $\gamma^\eta_i$, $\forall{\eta \in \{r, g, b\}}$ are drawn at random for each image $I_i$ with $i \in [1, N_{test}]$ independently.

The sample images generated using the proposed color channel perturbation (CCP) method in fixed and variable schemes are shown in the $2^{nd}$ and $3^{rd}$ rows, respectively in Fig. \ref{fig:sample} corresponding to the original images shown in the $1^{st}$ row of the same figure. The original sample images are taken from the CIFAR10 dataset \cite{cifar}. It can be easily seen that the semantic meaning of the images with respect to the underlying objects within the image is preserved in the generated images. Even then there is a significant performance drop in image classification due to the CCP transformation. The color distribution of the output images using fixed scheme is similar (see $2^{nd}$ row) due to the fixed random weights used for all images in this scheme. However, the different color distributions (see $3^{rd}$ row) can be observed in the images generated using variable random weights. The sample images generated using the Adversarial attack \cite{goodfellow2014explaining} and the One-pixel attack \cite{su2019one} are also shown in $4^{th}$ and $5^{th}$ rows, respectively. It can be observed that the adversarial and one-pixel attacks try to follow the similar color distributions as in the original images.

The histograms of Red, Green and Blue channels for sample images are illustrated in Fig. \ref{fig:hist} by red, green and blue colors, respectively. The sample images in the first three rows and the last three rows are taken from CIFAR10 \cite{cifar} and Caltech256 \cite{caltech256} datasets, respectively. The sample images in the $1^{st}$ column are the original images, whereas the images in the other columns are generated using the different transformation approaches. The $2^{nd}$ and $3^{rd}$ columns are corresponding to the proposed CCP attack based transformation under fixed random and variable random settings, respectively. The $4^{th}$ and $5^{th}$ columns are corresponding to the One-pixel attack \cite{su2019one} and Adversarial attack \cite{goodfellow2014explaining} based transformations, respectively.
The x-axis and y-axis in each plot represent the bin and frequency, respectively. 
It is noticed from these plots that the proposed CCP attack leads to the variations in the color distributions using both fixed and variable random weight schemes (i.e., 2nd and 3rd columns, respectively) preserving the density distribution of the intensity values. These properties of CCP transformation enforces the different color combinations in the generated image while retaining the visual appearance and semantic meaning of the objects.
The histogram plots generated using the One-pixel attack \cite{su2019one} ($4^{th}$ column) and Adversarial attack \cite{goodfellow2014explaining} ($5^{th}$ row) show that these transformations try to preserve only the density distribution without preserving the color variation.

\begin{figure*}[!t]
\centering
\includegraphics[trim=10 75 15 45, width=0.9\linewidth, clip]{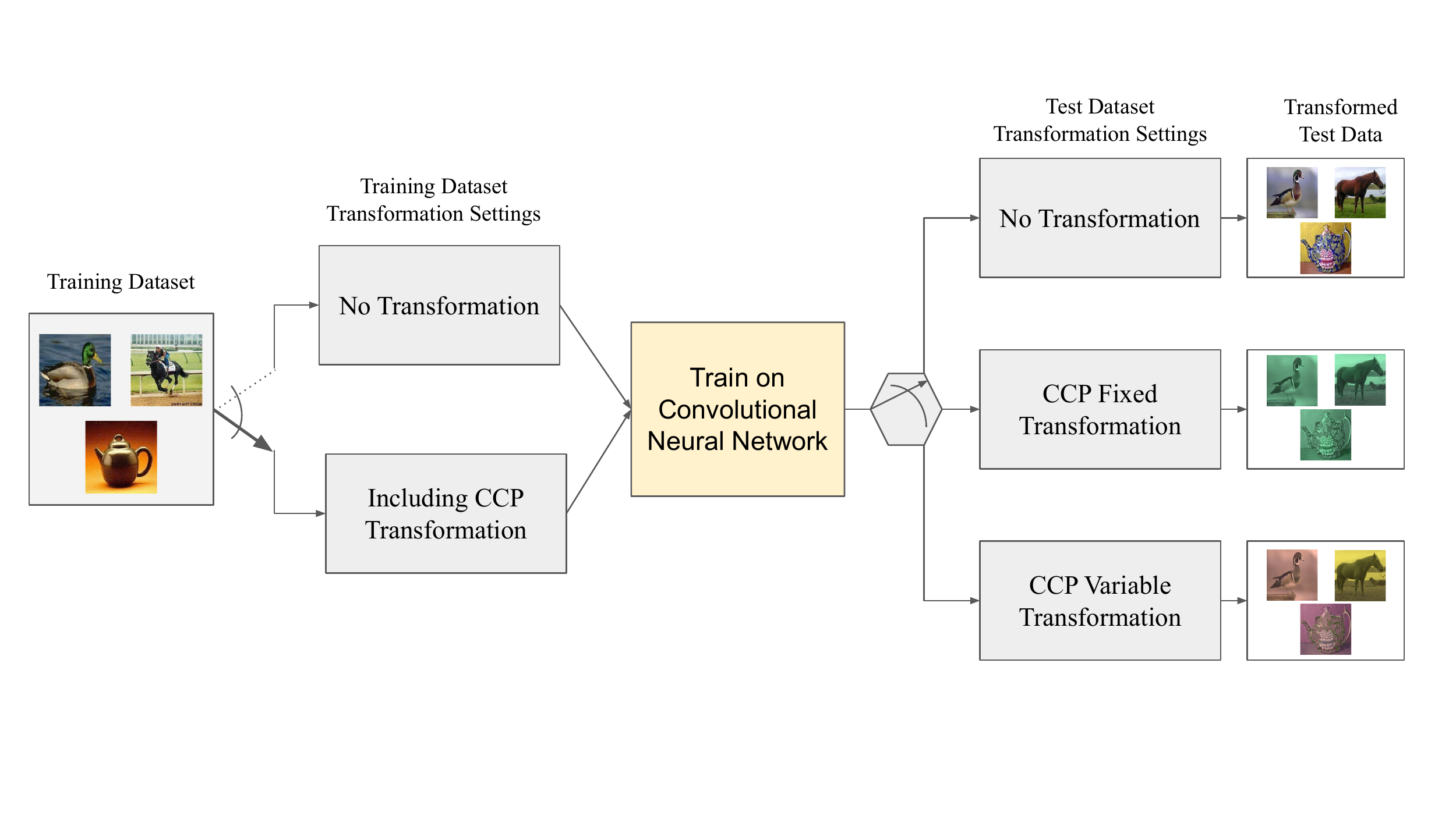}
\caption{Different experimental settings used in this work with the proposed CCP attack over training and test sets for the analysis of CNN fooling and robustness.}
\label{fig:ccp_settings}
\end{figure*}

We generate two transformed test sets using the proposed CCP transformation using the original test set as depicted in the right side in Fig. \ref{fig:ccp_settings}. The two transformed settings used are 1) CCP attack using fixed random weights for all the test examples (i.e., $CCP_f$) and 2) CCP attack using variable random weights for different test examples (i.e., $CCP_v$). The number of images in each transformed test set is same as the original test set. In order to show the impact and defense of CCP attack, we perform the experiments using two training sets, including 1) the original training set having training images without CCP based augmentation and 2) the modified training set having training images with CCP based augmentation as depicted on the left side of Fig. \ref{fig:ccp_settings}.

\section{Experimental Setup}
In this section, we describe the experimental setup used in the result analysis. First, we discuss about the CNN architectures, including VGG, ResNet and DenseNet. Then we elaborate the datasets, including CIFAR10, Caltech256 and TinyImageNet. Finally, we provide the details of settings for training.

\subsection{CNN Architectures Used}
In order to show the impact of the proposed CCP attack, the widely used CNN models such as VGG \cite{vgg}, ResNet \cite{resnet} and DenseNet \cite{densenet} are employed in the experiments. The VGG network is a deep CNN model with either 16 or 19 learnable layers. In the experiments, the VGG network with 16 layers (i.e., VGG16) is used over CIFAR10 and TinyImageNet datasets, whereas the VGG network with 19 layers (i.e., VGG19) is used over the Caltech256 dataset.
The ResNet architecture is built by arranging the residual blocks in a hierarchical order. A residual block transforms an input ($A$) to an output ($B$) as $B = A + h(A)$, where the residual function $h(.)$ consists of two convolution layer and an activation function. The residual model has shown very promising performance with respect to convergence in the training of the deep CNNs, which otherwise fail to converge as it provides the highway for gradient flow through identity mapping. In the experiments, the ResNet56, ResNet18 and ResNet101 models are used over CIFAR10, Caltech256 and TinyImageNet datasets, respectively.
We have also used the DenseNet network to observe the performance drop of heavy CNN using the proposed CCP attack. The DenseNet network contains the identity mapping from a layer to all the following layers, thus becomes a complex model. The DenseNet121 model has been used in the experiments over all the datasets.

\subsection{Datasets Used}
We perform the experiments under image classification framework over three benchmark datasets (i.e., CIFAR10 \cite{cifar}, Caltech256 \cite{caltech256} and TinyImageNet \cite{tinyimagenet}) to observe the effect of the proposed CCP attack. 
We consider the most popular CIFAR10 dataset\footnote{https://www.cs.toronto.edu/~kriz/cifar.html} \cite{cifar} which is widely being used to test the performance of CNNs for image classification. The CIFAR10 dataset consists of $60,000$ images from $10$ object categories with $6,000$ images per category. Out of $60,000$ images, $10,000$ images (i.e., $1,000$ images per class) are provided as the test set and remaining images are provided in the training set. Each transformed test set also contains $10,000$ images with $1,000$ images per class. 
As the CIFAR10 dataset contains the low resolution images (i.e., $32 \times 32$), we also use the datasets having better resolution images, like the TinyImageNet dataset having $64 \times 64$ resolution images and the Caltech256 dataset having $256 \times 256$ resolution images.
The Caltech256 dataset\footnote{http://www.vision.caltech.edu/Image\_Datasets/Caltech256/} \cite{caltech256} consists of $30,607$ images from $257$ object categories. The $80\%$ of the Caltech256 dataset is used for the training and remaining for the testing purpose. The overall complexity of the Caltech256 dataset is high due to the presence of the larger category sizes, larger category clutter and overall increased difficulty. 
The benchmark TinyImageNet dataset\footnote{https://tiny-imagenet.herokuapp.com/} \cite{tinyimagenet} is also included in the experiments. The TinyImageNet dataset is a subset of the original ImageNet large-scale visual recognition challenge \cite{russakovsky2015imagenet}. The TinyImageNet dataset contains $1,00,000$ images in the training set, $10,000$ images in the validation set, and $10,000$ images in the test set with $200$ object categories having $500$ training images, $50$ validation images and $50$ test images in each category.

\begin{figure}[!t]
\centering
\includegraphics[trim=110 50 148 18, width=\columnwidth, clip]{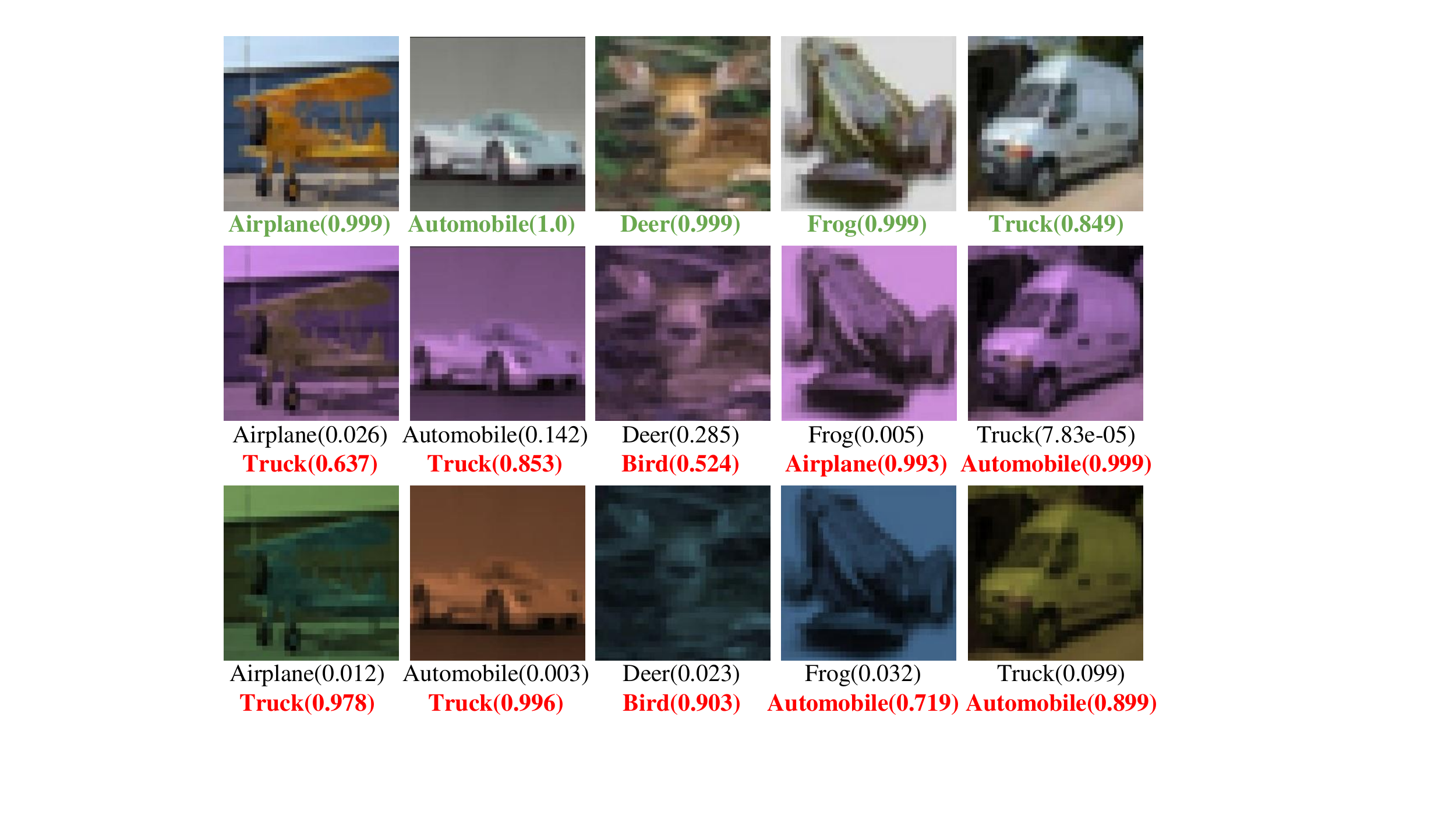}
\caption{The sample results depicting the effect of the proposed CCP transformations to fool the CNN using DenseNet model \cite{densenet}. The $1^{st}$ row shows the image classification results over the sample images from the original test set of CIFAR10 dataset \cite{cifar}. The $2^{nd}$ and $3^{rd}$ rows present the classification results of the same images after CCP attack with fixed and variable random weight settings, respectively. The class labels with probability for correct and misclassified classes are also shown. Best viewed in color.}
\label{fig:failed_cases_cifar}
\end{figure}

\subsection{Training Settings}
For all the experiments, the Keras framework with Tensorflow at the backend is used.
We have performed the training of ResNet56 over the original training set of CIFAR10 dataset. The model is trained for $200$ epochs with a batch size of $32$. The learning rate is set to $1e^{-3}$, $1e^{-4}$, $1e^{-5}$, $1e^{-6}$, and $0.5e^{-6}$ for $100$ epochs, $40$ epochs, $30$ epochs, $10$ epochs, and $20$ epochs, respectively. The Adam optimizer \cite{adam} is used with categorical cross-entropy loss function. Following data augmentations are applied during training: normalization with zero mean and unit standard deviation at dataset as well as image level, ZCA whitening with epsilon of $1e^{-06}$, random rotation from $0$ to $180$ degree, random shifting (range $0.1$) horizontally and vertically, random flipping horizontally and vertically, random shear, random zoom, and random channel shifts. 
The training of DenseNet121 is carried out over the original training set of CIFAR10 dataset. The model is trained for $85$ epochs with learning rate as $0.001$ for the first $50$ epochs, $0.0001$ for next $25$ epochs and $0.00001$ for the last $5$ epochs.  
The data augmentations, including random rotation (range $40$), height and width shifts, horizontal flips, and random zoom (range $0.1$), are used during the training.
We have also done training of VGG16 over CIFAR10 dataset with data augmentations such as feature and sample normalization, ZCA whitening, height and width shift, horizontal and vertical flip, and random rotation (range $15$). The learning rate ($lr$) is varied in each epoch as $lr = lr \times (0.5^{epoch/lr\_drop})$, where $lr\_drop$ is a factor used to reduce the learning rate.

The transfer learning is utilized for VGG19, ResNet18 and DenseNet121 over the Caltech256 dataset with Adam optimizer \cite{adam} using categorical cross-entropy loss function. The data augmentations used for this training setting are: random rotation from $0$ to $40$ degree, width and height shift range up to $0.2$, random shear, random zoom and horizontal flips. For Caltech256 dataset, the Adam optimizer is used with learning rate of $0.001$ and $0.0001$ for the first $20$ epochs and next $20$ epochs, respectively.

\begin{figure}[!t]
\centering
\includegraphics[trim=100 52 160 16, width=\columnwidth, clip]{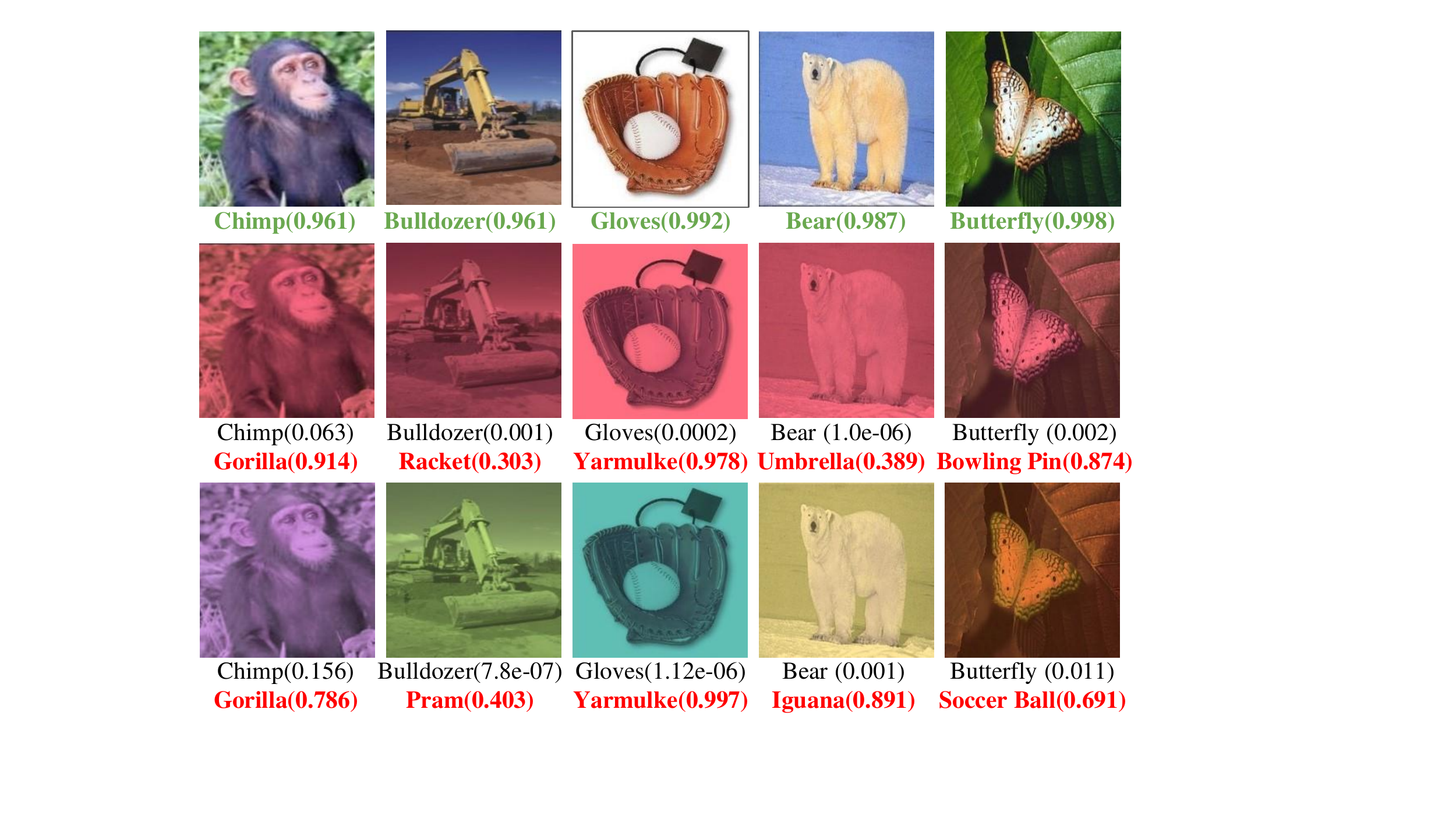}
\caption{The sample results depicting the effect of the proposed CCP transformations to fool the CNN using ResNet18 model \cite{resnet}. The $1^{st}$ row shows the image classification results over the sample images from the original test set of Caltech256 dataset \cite{caltech256}. The $2^{nd}$ and $3^{rd}$ rows present the classification results of the same images after CCP attack with fixed and variable random weight settings, respectively. The class labels with probability for correct and misclassified classes are also shown. Best viewed in color.}
\label{fig:failed_cases_caltech}
\end{figure}

The VGG16, ResNet101 and DenseNet121 are trained over TinyImageNet with data augmentations like Gaussian blur over $50\%$ of the images with random sigma between $0$ to $0.5$, horizontal and vertical flips, cropping images by $-10\%$ to $20\%$, and affine transformations such as scaling up to $80\%$ to $120\%$ of image/height, translation by $-20$ to $+20$ relative to height/width (per axis), rotation by $-45$ to $45$ degrees, shear by $-16$ to $16$ degrees, Coarse Dropout by $2-5\%$, and brightness and contrast normalization. For TinyImageNet dataset, the Adam optimizer is used with learning rate of $0.001$ along with cyclic learning rates of $0.0001 - 0.0006$, $0.00001 - 0.00006$ and $0.000001 - 0.000006$ for $48$ epochs, $12$ epochs and next $12$ epochs, respectively using the categorical cross-entropy loss function.

The training and test sets used for the experiments are summarized in Fig. \ref{fig:ccp_settings}. First, the experiments are performed over the original training images to show the impact of CCP attacks on test set images. Later, the CCP attacked training images are also used in the training set to increase the robustness of the model against the CCP attack on test set images. The settings for training are same for both types of images in the training set with or without the CCP attack. Note that the values of scale $s$ and bias $b$ in the CCP attack are set to $2$ and $0$, respectively, for the CIFAR10 dataset and $1$ and $30$, respectively, for the Caltech256 and TinyImagenet datasets for visually appealing generated images.

\section{Experiments, Results and Observations}
This section is devoted for the experimental results and analysis. First, we show the impact of the proposed CCP attack with qualitative and quantitative results. Then, we compare the results with the existing methods. Finally, we enhance the defense capability of the CNNs using CCP attack based augmentation of training data against such attacks.

\subsection{Qualitative Results}
The sample results depicting the effect of the proposed CCP transformations to fool the CNN are illustrated in Fig. \ref{fig:failed_cases_cifar} and Fig. \ref{fig:failed_cases_caltech}.
The DenseNet121 model is used over the sample images from CIFAR10 dataset in Fig. \ref{fig:failed_cases_cifar} and the ResNet18 model is used over the sample images from Caltech256 dataset in Fig. \ref{fig:failed_cases_caltech}.
The $1^{st}$ row, in both Fig. \ref{fig:failed_cases_cifar} and \ref{fig:failed_cases_caltech}, show the classification results using the images of the corresponding original test set. The $2^{nd}$ and $3^{rd}$ rows, in both Fig. \ref{fig:failed_cases_cifar} and \ref{fig:failed_cases_caltech}, present the classification results of the corresponding $1^{st}$ row images after CCP attack with fixed ($CCP_f$) and variable ($CCP_v$) random weights, respectively. The class labels along with the probability of classification for correct and misclassified classes are also shown. 
It can be observed in Fig. \ref{fig:failed_cases_cifar} that the original image classified in `Automobile' category with $100\%$ confidence gets completely misclassified in `Truck' category after CCP attack using fixed and variable random weights with $85.3\%$ and $99.6\%$ confidence, respectively. A very similar misclassification is reported for the other samples as well after CCP attack.
The similar trend can also be noticed in Fig. \ref{fig:failed_cases_caltech} over samples of Caltech256 dataset. The original image of category `Chimp' with $0.9608$ probability has been classified as `Gorilla' with $0.9136$ and $0.7857$ probabilities under fixed and variable settings, respectively. Similarly, the sample image from `Gloves' category with $0.9921$ probability is completely misclassified in `Yarmulke' category with $0.9784$ and $0.9965$ probabilities after $CCP_f$ and $CCP_v$ attacks, respectively. Moreover, all of the original input images have been misclassified in other classes with high confidence after CCP attack as shown in Fig. \ref{fig:failed_cases_cifar} and \ref{fig:failed_cases_caltech}. However, the visual appearance of the original images and attacked images are pretty similar with some color perturbation. It shows that the color plays an important role in decision making by CNN and the current training procedure is not able to exhibit the robustness of CNN for color channel perturbations.

\begin{table}[!t]
\centering
\caption{Experimental results in terms of the accuracy ($\%$) over \textbf{original CIFAR10 training data} and different test sets. Total 30 trials are performed. The $CCP_f$ and $CCP_v$ represent the results using the proposed CCP attack with fixed and variable random weight settings, respectively. The STD denotes the standard deviation. The minimum and maximum denotes the max and min accuracy out of 30 trials, respectively. The original refers to the accuracy over original test set.}
\begin{tabular}{m{0.13\columnwidth}m{0.089\columnwidth}m{0.089\columnwidth}m{0.089\columnwidth}m{0.089\columnwidth}m{0.089\columnwidth}m{0.089\columnwidth}}
\toprule
\multirow{2}{*}{\textbf{30 Trials}} &\multicolumn{2}{c}{\textbf{VGG16}}
&\multicolumn{2}{c}{\textbf{ResNet56}} &\multicolumn{2}{c}{\textbf{DenseNet121}}\\
\cmidrule{2-7}
&$CCP_f$ &$CCP_v$ &$CCP_f$ &$CCP_v$ &$CCP_f$ &$CCP_v$\\\midrule
\textbf{Mean} &\textbf{76.49} &\textbf{75.92} &\textbf{78.38} &\textbf{77.58} &\textbf{60.23} &\textbf{56.46}\\
\textbf{STD} &9.16 &0.29 &7.54 &0.29 &10.93 &0.35\\
\textbf{Minimum} &53.59 &75.20 &59.99 &76.94 &37.77 &55.68\\
\textbf{Maximum} &88.31 &76.36 &88.55 &78.34 &77.95 &56.99\\
\hline
\textbf{Original} &\multicolumn{2}{c}{\textbf{93.58}}
&\multicolumn{2}{c}{\textbf{91.44}} &\multicolumn{2}{c}{\textbf{92.72}}\\
\bottomrule
\end{tabular}
\label{table:results_cifar}
\end{table}

\begin{table}[!t]\centering
\caption{Experimental results in terms of the accuracy ($\%$) over \textbf{original Caltech256 training data} and different test sets. Thirty trials are performed. }
\begin{tabular}{m{0.13\columnwidth}m{0.089\columnwidth}m{0.089\columnwidth}m{0.089\columnwidth}m{0.089\columnwidth}m{0.089\columnwidth}m{0.089\columnwidth}}
\toprule
\multirow{2}{*}{\textbf{30 Trials}} &\multicolumn{2}{c}{\textbf{VGG19}} &\multicolumn{2}{c}{\textbf{ResNet18}}
&\multicolumn{2}{c}{\textbf{DenseNet121}}\\
\cmidrule{2-7}
&$CCP_f$ &$CCP_v$ &$CCP_f$ &$CCP_v$ &$CCP_f$ &$CCP_v$\\\midrule
\textbf{Mean} &\textbf{28.76} &\textbf{28.01} &\textbf{38.01} &\textbf{41.58} &\textbf{38.10} &\textbf{38.56}\\
\textbf{STD} &6.65 &0.36 &9.60 &0.43 &5.98 &2.25\\
\textbf{Minimum} &16.14 &27.29 &14.77 &40.89 &26.73 &32.67\\
\textbf{Maximum} &40.55 &28.80 &51.70 &42.61 &49.32 &42.51\\
\hline
\textbf{Original} &\multicolumn{2}{c}{\textbf{50.42}} &\multicolumn{2}{c}{\textbf{66.73}} &\multicolumn{2}{c}{\textbf{64.83}}\\
\bottomrule
\end{tabular}
\label{table:results_caltech}
\end{table}

\begin{table}[!t]\centering
\caption{Experimental results in terms of the accuracy ($\%$) over \textbf{original TinyImageNet training data} and different test sets. Thirty trials are performed. }
\begin{tabular}{m{0.13\columnwidth}m{0.089\columnwidth}m{0.089\columnwidth}m{0.089\columnwidth}m{0.089\columnwidth}m{0.089\columnwidth}m{0.089\columnwidth}}
\toprule
\multirow{2}{*}{\textbf{30 Trials}} &\multicolumn{2}{c}{\textbf{VGG16}}
&\multicolumn{2}{c}{\textbf{ResNet101}}
&\multicolumn{2}{c}{\textbf{DenseNet121}} \\
\cmidrule{2-7}
&$CCP_f$ &$CCP_v$ &$CCP_f$ &$CCP_v$ &$CCP_f$ &$CCP_v$\\\midrule
\textbf{Mean} &\textbf{13.31} &\textbf{14.07} &\textbf{26.09} &\textbf{26.72} &\textbf{30.55} &\textbf{31.09} \\
\textbf{STD} &4.04 &0.29 &4.81 &0.29 &4.09 &0.25\\
\textbf{Minimum} &3.34 &13.48 &12.95 &26.21 &18.26 &27.90 \\
\textbf{Maximum} &20.26 &14.70 &33.54 &27.38 &37.86 &29.23 \\
\hline
\textbf{Original} &\multicolumn{2}{c}{\textbf{42.55}}
&\multicolumn{2}{c}{\textbf{59.19}} &\multicolumn{2}{c}{\textbf{62.52}} \\
\bottomrule
\end{tabular}
\label{table:results_tinyimagenet}
\end{table}

\begin{table*}[!t]
\caption{The comparison of results for different attacks on the original CIFAR10, Caltech256 and TinyImageNet test sets using VGG, ResNet and DenseNet Models. The results of the proposed CCP attack are reported under both fixed and variable random weight settings. These results are computed as an average and standard deviation over $30$ trials. The impact of the attack is also mentioned in terms of the accuracy drop in \% w.r.t. the accuracy without attack. The best and second best improvements are highlighted in Bold and Italic, respectively. The $\uparrow$ and $\downarrow$ represent the gain and loss, respectively.}
\centering
\begin{tabular}{m{4cm}m{4cm}m{4cm}m{4cm}}
\hline
\multicolumn{4}{c}{\textbf{Accuracy over CIFAR10 Dataset}}\\
\hline
\textbf{Type of Attack} & \textbf{VGG16} & \textbf{ResNet56} & \textbf{DenseNet121}  \tabularnewline
\hline
Without Attack & $93.58$ $\pm$ $0.00$ & $91.44$ $\pm$  $0.00$ & $92.72$ $\pm$  $0.00$ \tabularnewline
One Pixel Attack& $92.81$ $\pm$ $0.10$ ($\downarrow$ $0.82\%$) & $89.40$  $\pm$ $0.16$ ($\downarrow$ $2.23\%$) & $92.21$ $\pm$  $0.07$ ($\downarrow$ $0.55\%$) \tabularnewline
Adversarial Attack & $82.61$ $\pm$  $0.21$ ($\downarrow$ $11.72\%$) & $79.78$ $\pm$ $0.79$ ($\downarrow$ $12.75\%$) & $67.43$ $\pm$ $0.21$ ($\downarrow$ $27.28\%$) \tabularnewline
CCP Attack ($CCP_f$) & $\textit{76.49}$ $\pm$ $\textit{9.16}$ ($\downarrow$ $\textit{18.26\%}$) & $\textit{78.38}$ $\pm$ $\textit{7.54}$ ($\downarrow$ $\textit{14.28\%}$) & $\textit{60.23}$ $\pm$ $\textit{10.9}$ ($\downarrow$ $\textit{35.04\%}$) \tabularnewline
CCP Attack ($CCP_v$) & $\textbf{75.92}$ $\pm$ $\textbf{0.29}$ ($\downarrow$ $\textbf{18.87\%}$) & $\textbf{77.58}$ $\pm$ $\textbf{0.29}$ ($\downarrow$ $\textbf{15.16\%}$) & $\textbf{56.46}$ $\pm$ $\textbf{0.35}$ ($\downarrow$ $\textbf{39.11\%}$) \tabularnewline
\hline
\multicolumn{4}{c}{\textbf{Accuracy over Caltech256 Dataset}}\\
\hline
\textbf{Type of Attack} & \textbf{VGG19} & \textbf{ResNet18} & \textbf{DenseNet121}  \tabularnewline
\hline
Without Attack & $50.42$ $\pm$  $0.00$ & $66.73$ $\pm$  $0.00$ & $64.83$ $\pm$  $0.00$ \tabularnewline
One Pixel Attack& $50.43$ $\pm$  $0.02$  ($\uparrow$ $0.02\%$) & $66.64$  $\pm$ $0.05$ ($\downarrow$ $0.13\%$) & $64.79$ $\pm$  $0.03$ ($\downarrow$ $0.06\%$) \tabularnewline
Adversarial Attack & $48.11$ $\pm$  $0.11$ ($\downarrow$ $4.58\%$) & $64.69$ $\pm$ $0.13$ ($\downarrow$ $3.06\%$) & $63.58$ $\pm$  $0.13$ ($\downarrow$ $1.93\%$) \tabularnewline
CCP Attack ($CCP_f$) & $\textit{28.76}$ $\pm$  $\textit{6.65}$ ($\downarrow$ $\textit{42.96\%}$) & $\textbf{38.01}$ $\pm$ $\textbf{9.60}$ ($\downarrow$ $\textbf{43.04\%}$) & $\textbf{38.10}$ $\pm$  $\textbf{5.98}$ ($\downarrow$ $\textbf{41.23\%}$) \tabularnewline
CCP Attack ($CCP_v$) & $\textbf{28.01}$ $\pm$  $\textbf{0.36}$ ($\downarrow$ $\textbf{44.45\%}$) & $\textit{41.58}$ $\pm$ $\textit{0.43}$ ($\downarrow$ $\textit{37.69\%}$) & $\textit{38.56}$ $\pm$  $\textit{2.25}$ ($\downarrow$ $\textit{40.52\%}$) \tabularnewline
\hline
\multicolumn{4}{c}{\textbf{Accuracy over TinyImageNet Dataset}}\\
\hline
\textbf{Type of Attack} & \textbf{VGG16} & \textbf{ResNet101} & \textbf{DenseNet121}  \tabularnewline
\hline
Without Attack & $42.55$ $\pm$ $0.00$ & $59.19$ $\pm$ $0.00$ & $62.52$ $\pm$ $0.00$ \tabularnewline
One Pixel Attack & $42.61$ $\pm$ $0.09$ ($\uparrow$ $0.14\%$) & $59.01$ $\pm$ $0.08$ ($\downarrow$ $0.30\%$) & $62.44$ $\pm$ $0.08$ ($\downarrow$ $1.28\%$) \tabularnewline
Adversarial Attack & $40.26$ $\pm$ $0.19$ ($\downarrow$ $5.38\%$) & $56.44$ $\pm$ $0.14$ ($\downarrow$ $4.65\%$) & $60.01$ $\pm$ $0.16$ ($\downarrow$ $4.01\%$) \tabularnewline
CCP Attack ($CCP_f$) & $\textbf{13.31}$ $\pm$ $\textbf{4.04}$ ($\downarrow$ $\textbf{68.72\%}$) & $\textbf{26.09}$ $\pm$ $\textbf{4.81}$ ($\downarrow$ $\textbf{55.92\%}$) & $\textbf{30.55}$ $\pm$ $\textbf{4.09}$ ($\downarrow$ $\textbf{51.14\%}$) \tabularnewline
CCP Attack ($CCP_v$) & $\textit{14.07}$ $\pm$ $\textit{0.29}$ ($\downarrow$ $\textit{66.93\%}$) & $\textit{26.72}$ $\pm$ $\textit{0.29}$ ($\downarrow$ $\textit{54.86\%}$) & $\textit{31.09}$ $\pm$ $\textit{0.25}$ ($\downarrow$ $\textit{50.27\%}$) \tabularnewline
\hline
\end{tabular}
\label{table:results_comparison}
\end{table*}

\subsection{Experimental Results with Original Training Data}
In this experiment, the VGG16, ResNet56 and DenseNet121 models are trained over the original CIFAR10 training set. The performance of trained models is tested over the original CIFAR10 test set as well as the transformed test sets using the proposed color channel perturbation (CCP) attack with fixed and variable random weight settings. We perform thirty trials of the experiment as the weights are generated randomly between $0$ and $1$.
The average classification test accuracy is reported in Table \ref{table:results_cifar} over different transformed test sets for the models trained over the original training data. The accuracy over original test is also reported in Table \ref{table:results_cifar}. The performance in different trials varies in $CCP_f$ and $CCP_v$ attack based transformed test sets due to the random weights used in their generation. It can be noted that the average performance degrades significantly after fixed and random weight based CCP attack by $17.09\%$ and $17.66\%$, respectively, using VGG16; $13.06\%$ and $13.86\%$, respectively, using ResNet56; and $32.49\%$ and $36.26\%$, respectively, using DenseNet121. It is also observed that the variable weight generation scheme leads to the lower standard deviation in the results as compared to the fixed weight generation scheme. It is due to the fact that the random weights for all the images are generated independently in each trial of variable setting. Whereas, in a trial of the fixed setting the random weights are same for all the images, thus leading to the high standard deviation in the results.

\begin{table}[!t]\centering
\caption{Experimental results in terms of the accuracy ($\%$) over \textbf{CCP augmented CIFAR10 training data} and different test sets. Thirty trials are performed.}
\begin{tabular}{m{0.13\columnwidth}m{0.089\columnwidth}m{0.089\columnwidth}m{0.089\columnwidth}m{0.089\columnwidth}m{0.089\columnwidth}m{0.089\columnwidth}}
\toprule
\multirow{2}{*}{\textbf{30 Trials}} &\multicolumn{2}{c}{\textbf{VGG16}}
&\multicolumn{2}{c}{\textbf{ResNet56}} &\multicolumn{2}{c}{\textbf{DenseNet121}}\\
\cmidrule{2-7}
&$CCP_f$ &$CCP_v$ &$CCP_f$ &$CCP_v$ &$CCP_f$ &$CCP_v$\\\midrule
\textbf{Mean} &\textbf{90.81} &\textbf{90.89} &\textbf{89.00} &\textbf{88.92} &\textbf{88.83} &\textbf{88.67} \\
\textbf{STD} &0.59 &0.61 &0.35 &0.67 &0.35 &0.12 \\
\textbf{Minimum} &89.16 &88.09 &88.31 &86.63 &88.23 &88.39 \\
\textbf{Maximum} &91.35 &91.43 &89.67 &89.59 &89.48 &88.95 \\
\hline
\textbf{Original} &\multicolumn{2}{c}{\textbf{91.42}} &\multicolumn{2}{c}{\textbf{90.15}} &\multicolumn{2}{c}{\textbf{90.61}} \\
\bottomrule
\end{tabular}
\label{table:results_cifar-augment}
\end{table}

The experimental results using ResNet18, VGG19 and DenseNet121 are summarized in Table \ref{table:results_caltech} over the Caltech256 dataset. In this experiment, the CNN models are trained over original training set and tested over original as well as different transformed test sets. We perform 30 trials over transformed test sets and present the average classification accuracy with standard deviation (STD) in Table \ref{table:results_caltech}. A very similar performance degradation is observed due to the proposed CCP attack over Caltech256 dataset also, in spite of being a high resolution dataset.
After applying the CCP transformation with variable random weight setting, the original accuracy of $50.42\%$, $66.73\%$ and $64.83\%$ using VGG19, ResNet18 and DenseNet121 models gets dropped to $28.01\%$, $41.58\%$ and $38.56\%$, respectively. It can also be seen that the standard deviation is lower in variable random weight setting as the weights are generated independently for each image. Note that the impact of the proposed CCP attack is also significant over the high resolution Caltech256 dataset.

\begin{table}[!t]\centering
\caption{Experimental results in terms of the accuracy ($\%$) over \textbf{CCP augmented Caltech256 training data} and different test sets. Thirty trials are performed.}
\begin{tabular}{m{0.13\columnwidth}m{0.089\columnwidth}m{0.089\columnwidth}m{0.089\columnwidth}m{0.089\columnwidth}m{0.089\columnwidth}m{0.089\columnwidth}}
\toprule
\multirow{2}{*}{\textbf{30 Trials}} &\multicolumn{2}{c}{\textbf{VGG19}} &\multicolumn{2}{c}{\textbf{ResNet18}}
&\multicolumn{2}{c}{\textbf{DenseNet121}}\\
\cmidrule{2-7}
&$CCP_f$ &$CCP_v$ &$CCP_f$ &$CCP_v$ &$CCP_f$ &$CCP_v$\\\midrule
\textbf{Mean} &\textbf{40.14} &\textbf{40.07} &\textbf{63.28} &\textbf{63.36} &\textbf{60.93} &\textbf{60.64} \\
\textbf{STD} &3.51 &0.25 &0.72 &0.24 &1.54 &0.16 \\
\textbf{Minimum} &31.00 &39.63 &60.72 &62.73 &57.28 &60.35 \\
\textbf{Maximum} &45.18 &40.68 &64.16 &63.69 &62.78 &61.11 \\
\hline
\textbf{Original} &\multicolumn{2}{c}{\textbf{52.57}} &\multicolumn{2}{c}{\textbf{64.87}} &\multicolumn{2}{c}{\textbf{62.58}} \\
\bottomrule
\end{tabular}
\label{table:results_caltech-augment}
\end{table}

\begin{table}[!t]\centering
\caption{Experimental results in terms of the accuracy ($\%$) over \textbf{CCP augmented TinyImageNet training data} and different test sets. Thirty trials are performed.}
\begin{tabular}{m{0.13\columnwidth}m{0.089\columnwidth}m{0.089\columnwidth}m{0.089\columnwidth}m{0.089\columnwidth}m{0.089\columnwidth}m{0.089\columnwidth}}
\toprule
\multirow{2}{*}{\textbf{30 Trials}} &\multicolumn{2}{c}{\textbf{VGG16}} &\multicolumn{2}{c}{\textbf{ResNet101}}
&\multicolumn{2}{c}{\textbf{DenseNet121}}\\
\cmidrule{2-7}
&$CCP_f$ &$CCP_v$ &$CCP_f$ &$CCP_v$ &$CCP_f$ &$CCP_v$\\\midrule
\textbf{Mean} &\textbf{34.02} &\textbf{33.82} &\textbf{51.87} &\textbf{51.84} &\textbf{55.64} &\textbf{55.49} \\
\textbf{STD} &1.09 &0.23 &1.05 &0.25 &1.02 &0.23 \\
\textbf{Minimum} &31.90 &33.43 &49.19 &51.38 &53.33 &54.98 \\
\textbf{Maximum} &35.99 &34.47 &53.37 &52.35 &57.21 &55.92 \\
\hline
\textbf{Original} &\multicolumn{2}{c}{\textbf{43.73}} &\multicolumn{2}{c}{\textbf{59.65}} &\multicolumn{2}{c}{\textbf{63.41}} \\
\bottomrule
\end{tabular}
\label{table:results_tinyimagenet-augment}
\end{table}

\begin{table*}[!t]
\caption{The comparison of results for different attacks on the original CIFAR10, Caltech256 and TinyImageNet test sets using VGG, ResNet and DenseNet Models augmented with CCP transformation in the training data. The results of the proposed CCP attack are reported under both fixed and variable random weight settings. These results are computed as an average and standard deviation over $30$ trials. The \% improvement in accuracy is also mentioned after CCP augmentation of training data w.r.t. the accuracy without training augmentation as presented in Table \ref{table:results_comparison}. The best and second best improvements are highlighted in Bold and Italic, respectively. The $\uparrow$ and $\downarrow$ represent the gain and loss, respectively.}
\centering
\begin{tabular}{m{4cm}m{4cm}m{4cm}m{4cm}}
\hline
\multicolumn{4}{c}{\textbf{Accuracy over CIFAR10 Dataset}}\\
\hline
\textbf{Type of Attack} & \textbf{VGG16} & \textbf{ResNet56} & \textbf{DenseNet121}  \tabularnewline
\hline
Without Attack & $91.42$ $\pm$  $0.00$ ($\downarrow$ $2.13\%$) & $90.15$ $\pm$  $0.00$ ($\downarrow$ $1.41\%$) & $90.61$ $\pm$  $0.00$ ($\downarrow$ $2.28\%$)\tabularnewline
One Pixel Attack& $90.57$ $\pm$  $0.08$ ($\downarrow$ $2.41\%$)& $89.07$  $\pm$ $0.17$ ($\downarrow$ $0.37\%$)& $89.12$ $\pm$  $0.12$ ($\downarrow$ $3.35\%$)\tabularnewline
Adversarial Attack & $81.36$ $\pm$  $0.18$ ($\downarrow$ $1.51\%$)& $80.62$ $\pm$ $0.61$ ($\uparrow$ $1.05\%$)& $75.08$ $\pm$ $0.21$ ($\uparrow$ $11.35\%$)\tabularnewline
CCP Attack ($CCP_f$) & $90.81$ $\pm$ $0.59$ ($\uparrow$ $\textit{18.72\%}$)& $89.00$ $\pm$ $0.35$ ($\uparrow$ $\textit{13.55\%}$)&$88.83$ $\pm$ $0.35$ ($\uparrow$ $\textit{47.48\%}$)\tabularnewline
CCP Attack ($CCP_v$) & $90.89$ $\pm$ $0.61$ (\textbf{$\uparrow$} $\textbf{19.72\%}$)& $88.92$ $\pm$ $0.67$ ($\uparrow$ $\textbf{14.62\%}$)& $88.67$ $\pm$ $0.12$ ($\uparrow$ $\textbf{57.05\%}$)\tabularnewline
\hline
\multicolumn{4}{c}{\textbf{Accuracy over Caltech256 Dataset}}\\
\hline
\textbf{Type of Attack} & \textbf{VGG19} & \textbf{ResNet18} & \textbf{DenseNet121}  \tabularnewline
\hline
Without Attack & $52.57$ $\pm$  $0.00$ ($\uparrow$ $4.26\%$)& $64.87$ $\pm$  $0.00$ ($\downarrow$ $2.79\%$)& $63.41$ $\pm$  $0.00$ ($\downarrow$ $2.19\%$)\tabularnewline
One Pixel Attack& $52.33$ $\pm$  $0.03$  ($\uparrow$ $3.77\%$)& $64.82$  $\pm$ $0.05$ ($\downarrow$ $2.73\%$)& $62.49$ $\pm$  $0.04$ ($\downarrow$ $3.55\%$)\tabularnewline
Adversarial Attack & $49.55$ $\pm$  $0.11$ ($\uparrow$ $2.99\%$)& $63.04$ $\pm$ $0.15$ ($\downarrow$ $2.55\%$)& $60.36$ $\pm$  $0.14$ ($\downarrow$ $5.06\%$)\tabularnewline
CCP Attack ($CCP_f$) & $40.14$ $\pm$  $3.51$ ($\uparrow$ $\textit{39.57\%}$)& $63.28$  $\pm$ $0.72$ ($\uparrow$ $\textbf{66.48\%}$)& $60.93$ $\pm$  $1.54$ ($\uparrow$ $\textbf{59.92\%}$)\tabularnewline
CCP Attack ($CCP_v$) & $40.07$ $\pm$  $0.25$ ($\uparrow$ $\textbf{43.06\%}$)& $63.36$ $\pm$ $0.24$ ($\uparrow$ $\textit{52.38\%}$)& $60.64$ $\pm$  $0.16$ ($\uparrow$ $\textit{57.26\%}$)\tabularnewline
\hline
\multicolumn{4}{c}{\textbf{Accuracy over TinyImageNet Dataset}}\\
\hline
\textbf{Type of Attack} & \textbf{VGG16} & \textbf{ResNet101} & \textbf{DenseNet121}  \tabularnewline
\hline
Without Attack & $43.73$ $\pm$ $0.00$ ($\uparrow$ $2.77\%$)& $59.65$ $\pm$ $0.00$ ($\uparrow$ $0.78\%$)& $63.41$ $\pm$ $0.00$ ($\uparrow$ $1.42\%$)\tabularnewline
One Pixel Attack & $43.55$ $\pm$ $0.09$ ($\uparrow$ $2.21\%$)& $59.59$ $\pm$ $0.09$ ($\uparrow$ $0.98\%$)& $63.14$ $\pm$ $0.11$ ($\uparrow$ $1.12\%$)\tabularnewline
Adversarial Attack & $40.87$ $\pm$ $0.14$ ($\uparrow$ $1.52\%$)& $56.88$ $\pm$ $0.19$ ($\uparrow$ $0.78\%$)& $60.53$ $\pm$ $0.18$ ($\uparrow$ $0.87\%$)\tabularnewline
CCP Attack ($CCP_f$) & \textbf{$34.02$} $\pm$ \textbf{$1.09$} ($\uparrow$ $\textbf{155.60\%}$)& $51.87$ $\pm$ $1.05$ ($\uparrow$ $\textbf{98.81\%}$)& $55.64$ $\pm$ $1.02$ ($\uparrow$ $\textbf{82.13\%}$) \tabularnewline
CCP Attack ($CCP_v$) & $33.82$ $\pm$ $0.23$ ($\uparrow$ $\textit{140.37\%}$)& $51.84$ $\pm$ $0.25$ ($\uparrow$ $\textit{94.01\%}$)& $55.49$ $\pm$ $0.23$ ($\uparrow$ $\textit{78.48\%}$)\tabularnewline
\hline
\end{tabular}
\label{table:results_comparison1}
\end{table*}

We also conduct experiments over TinyImageNet dataset using VGG16, ResNet101, and DenseNet121 models and report the results for 30 trials of transformed sets and original test set in Table \ref{table:results_tinyimagenet}. A similar performance degradation can also be observed using all the models over TinyImageNet dataset. The performance of VGG16, ResNet101, and DenseNet121 is $42.55\%$, $59.19\%$, and $62.52\%$, respectively, over the original test set. The performance of the same networks gets reduced to $14.07\%$, $26.72\%$ and $31.09\%$, respectively, over the CCP attacked test set.

The results obtained over CIFAR10, Caltech256 and TinyImageNet datasets using VGG, ResNet and DenseNet models show the high impact on the performance drop of the CNNs using the CCP attack. Moreover, the CPP attack fools the CNNs, irrespective of the image resolution and CNN model. The accuracy obtained over different dataset using different model in each trial is provided as supplementary result.

\subsection{Results Comparison with Other Attacks}
We also compare the results of the proposed CCP attack with the existing attacks, such as One-pixel attack \cite{su2019one} and Adversarial attack \cite{goodfellow2014explaining}. The untargeted One-pixel attack is easy to perform as it has the impact over selected pixels. However, the adversarial attack involves the addition of the designed noise which is a complex process.
The comparison results are reported in Table \ref{table:results_comparison} with respect to the classification accuracy. The \% accuracy drop is also mentioned in the results for all the attacks w.r.t. the accuracy without attack. It is noticed that the proposed CCP attacks fool the CNN greatly as compared to the other attacks. The One-pixel attack exhibits the limited impact in terms of fooling the CNN. However, the Adversarial attack is close to the proposed CCP attack. The CCP attack is not data dependent in contrast to the Adversarial attack. Moreover, the CCP attack is much simpler to perform as compared to the Adversarial attack. Moreover, the proposed CCP attack is the first of its kind to exhibit the high performance degradation of the CNNs by simple color perturbations.

\subsection{Experimental Results with CCP Augmented Training Data}
We also perform the experiments by including the CCP transformed images in the training datasets. The variable random weight setting is used to generate the new training images using the proposed CCP transformation. However, the test sets remain same as in the earlier experiments. The experimental results using different CNN models over the augmented CIFAR10, Caltech256 and TinyImageNet datasets are reported in Table \ref{table:results_cifar-augment}, \ref{table:results_caltech-augment} and \ref{table:results_tinyimagenet-augment}, respectively. 
The testing is performed over the original test set and CCP attacked test sets. The testing is done for 30 trials. The average and the standard deviation are shown in the results.
It can be seen from these results that the robustness of the trained models increase greatly against the CCP attack after adding the CCP transformed images in the training set as compared to the results observed without CCP based augmentation illustrated in Section IV.B.
The improvements average performance over 30 trials in \% using VGG16, ResNet56 and DenseNet121 models over CIFAR10 dataset (Table \ref{table:results_cifar} vs Table \ref{table:results_cifar-augment}) are $\{14.32, 14.97\}$, $\{10.62, 11.34\}$ and $\{28.60, 32.21\}$ with $\{CCP_f, CCP_v\}$ test sets, respectively, after augmentation in the training set.
A slight drop in the performance over the original test set is also noticed using VGG16, ResNet56 and DenseNet121 models as $93.58 \rightarrow 91.42$, $91.44 \rightarrow 90.15$ and $92.72 \rightarrow 90.61$, respectively. It is due to the increased generalization of the models for CCP attacked images. However, the improvement over $CCP_f$ and $CCP_v$ test sets is more significant than the little drop over original test set. The similar trend is also observed using different models over the Caltech256 (Table \ref{table:results_caltech} vs Table \ref{table:results_caltech-augment}) and TinyImageNet (Table \ref{table:results_tinyimagenet} vs Table \ref{table:results_tinyimagenet-augment}) datasets. These results show that augmenting the training set with CCP transformed images increases the defense capability of networks against such attacks over the test data.

\subsection{Results Comparison under CCP Augmented Training Data}
The comparison results for different attacks over CIFAR10, Caltech256 and TinyImageNet datasets are summarized in Table \ref{table:results_comparison1} using different CNN models, when the CCP based augmentation is used during training. 
The improvement in the accuracy with CCP augmentation over the accuracy without CCP augmentation (i.e., w.r.t. Table \ref{table:results_comparison}) is also computed and shown in Table \ref{table:results_comparison1}. 
It can be noticed that the accuracies for CCP attack on test data are increased significantly, when CCP augmentation is done across all the datasets during the training of all the models. The highest performance gained by VGG16 is $155.60\%$ for CCP attack over the TinyImageNet dataset. Moreover, the accuracy for the original test set and other attacks are improved marginally over the TinyImageNet dataset. The performance for all the test sets is also improved with VGG19 over Caltech dataset. The performance under adversarial attack using CIFAR10 dataset over DenseNet121 is increased by $11.35\%$.

We also test the performance of different attacks under adversarial attack based augmentation during the training (these results are provided in supplementary). It is revealed that under adversarial data augmentation the performance of adversarial attack improves, whereas the performance of CCP attacks further deteriorates under adversarial attack based augmentation. It shows the suitability of CCP based augmentation against such attacks.
It also opens the future research direction to develop the multi-attack representative data augmentation to increase the defense capability of the CNN models.

\section{Conclusion}
A color channel perturbation (CCP) attack is proposed in this paper to fool the CNN for image classification. The proposed CCP attack is a simple attack and data independent. It uses the Red, Green and Blue channels of the original image to generate the Red, Green and Blue channels of the transformed image. Two transformed test sets are generated based on the fixed and variable random weight schemes. A significant drop is observed in the classification performance over the transformed test sets as compared to the original test set when used with the VGG, ResNet, and DenseNet models over the CIFAR10, Caltech256, and TinyImageNet datasets. The original samples correctly classified with high confidence are incorrectly classified with high confidence after CCP attack. The promising impact of the attack is observed using the proposed CCP attacks as compared to the state-of-the-art One-pixel and Adversarial attacks with respect to the accuracy drop. 
It is also observed from the histogram analysis that the proposed CCP attack preserves the density distribution, but changes the color contribution. However, the semantic meaning of images generated after the CCP attack is preserved. The proposed CCP attack is also used over the training data to observe the effect of such images over training to enhance the defense capability of CNN models. It is observed that the robustness of the model increases with the CCP transformed images in the training set. The experimental results confirm the high impact of the proposed CCP attack in fooling the CNN in image classification. Results also confirm the enhanced defense capability of the CNN models trained with the augmented training data with such attacks.

\section*{Acknowledgment}
The authors would like to thank Google for providing the Colaboratory service for the GPU accecerated computation used to compute the results in this paper.

\bibliographystyle{IEEEtran}
\bibliography{references}

\begin{IEEEbiography}[{\includegraphics[width=1in,height=1.25in,clip,keepaspectratio]{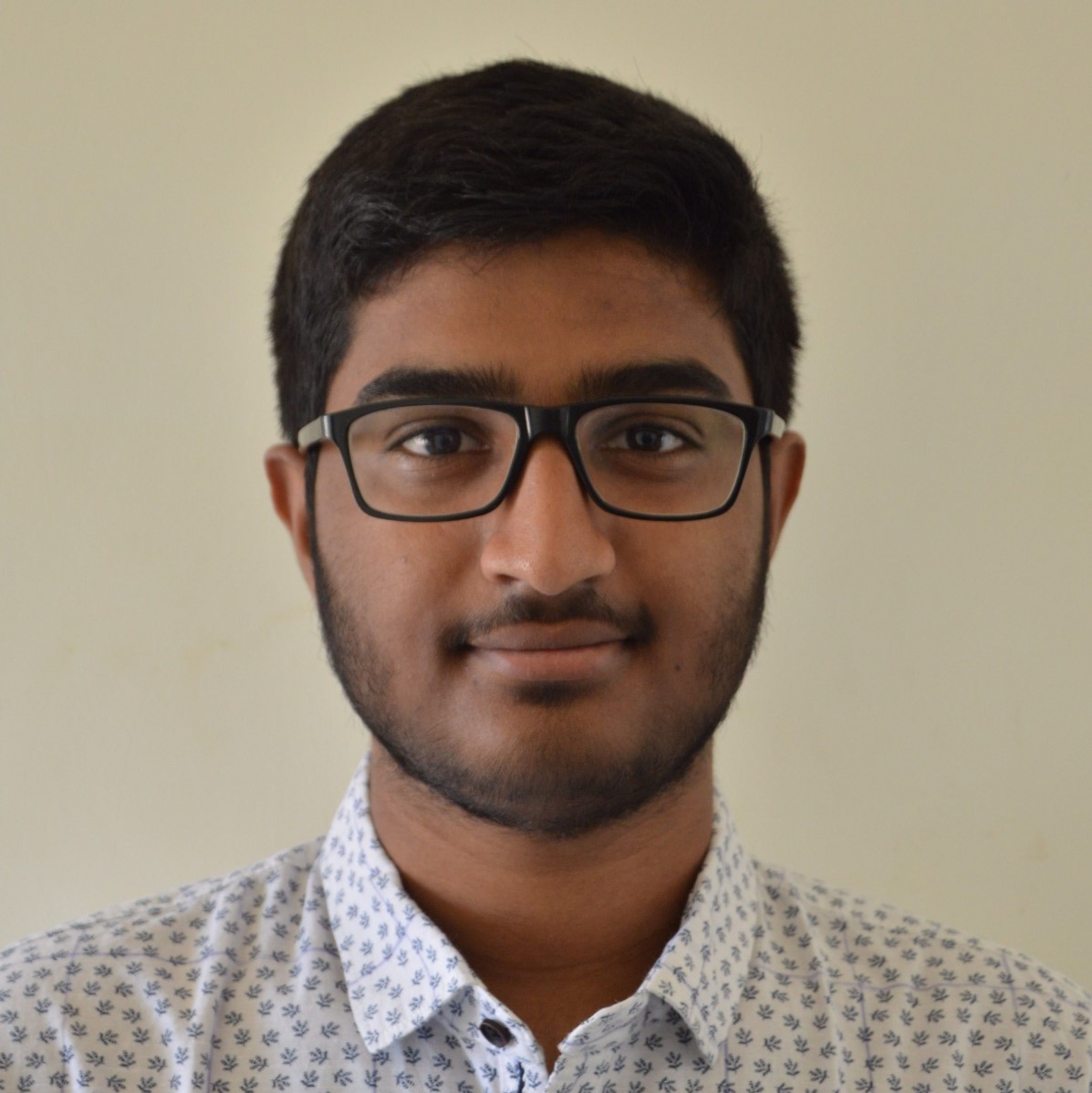}}]{Jayendra Kantipudi}{\space} was born in Hyderabad, Telangana, India in 1999. He is an undergraduate in Bachelors of Technology at the Indian Institute of Information Technology (IIIT) Sri City, Chittoor majoring in Computer Science and Engineering. He has completed his pre-college education in Hyderabad, Telangana, India.

He has worked as a Computer Vision Intern at ALOG TECH, a startup based company in Hyderabad, Telangana, India. He has worked as a Teaching Assistant in courses Database Management Systems and Signals and systems at IIIT Sri City from 2019 to 2020. His research interests include computer vision and data science. 
\end{IEEEbiography}

\begin{IEEEbiography}[{\includegraphics[width=1in,height=1.25in,clip,keepaspectratio]{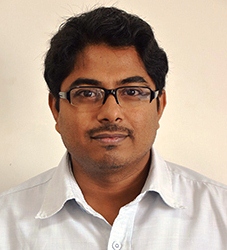}}]{Shiv Ram Dubey}{\space}(M'14){\space} has been with the Indian Institute of Information Technology (IIIT), Sri City since June 2016, where he is currently the Assistant Professor of Computer Science and Engineering. He received the Ph.D. degree in Computer Vision and Image Processing from Indian Institute of Information Technology, Allahabad (IIIT Allahabad) in 2016. Before that, from August 2012-Feb 2013, he was a Project Officer in the Computer Science and Engineering Department at Indian Institute of Technology, Madras (IIT Madras). 

He was a recipient of several awards including the Indo-Taiwan Joint Research Grant from DST/GITA, Govt. of India, Best PhD Award in PhD Symposium, IEEE-CICT2017 at IIITM Gwalior, Early Career Research Award from SERB, Govt. of India and NVIDIA GPU Grant Award Twice from NVIDIA. He received Outstanding Certificate of Reviewing Award from Information Fusion, Elsevier in 2018. He also received the Best Paper Award in IEEE UPCON 2015, a prestigious conference of IEEE UP Section. 

His research interest includes Computer Vision, Deep Learning, Image Processing, Biometrics, Medical Imaging, Convolutional Neural Networks, Image Feature Description, Content Based Image Retrieval, Image-to-Image Transformation, Face Detection and Recognition, Facial Expression Recognition, Texture and Hyperspectral Image Analysis. 
\end{IEEEbiography}

\begin{IEEEbiography}[{\includegraphics[width=1in,height=1.25in,clip,keepaspectratio]{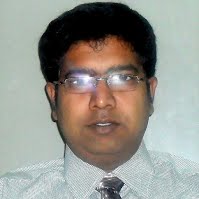}}]{Soumendu Chakraborty}{\space} is currently working as an Assistant Professor at Indian Institute of Information Technology (IIIT), Lucknow, U.P., India. He received his Bachelor of Engineering (B.E.) in Information Technology from University Institute of Technology, University of Burdwan, India in 2005. He did his M.Tech. in Computer Science and Engineering from GLA University, Mathura, India in 2013. He received his Ph.D. from Indian Institute of Information Technology, Allahabad, U.P., India in the year 2018.  He has 12 years of teaching and research experience. His research interests include face recognition and retrieval, image processing, biometric systems, image stegnography and pattern recognition.
\end{IEEEbiography}

\onecolumn
\section*{\huge{\bf Supplementary Material}} 
\vspace{20pt}

\noindent\rule{\linewidth}{0.4pt}
\noindent \textbf{Paper Title:} Color Channel Perturbation Attacks for Fooling Convolutional Neural Networks and A Defense Against Such Attacks\\
\textbf{Authors:} Jayendra Kantipudi, Shiv Ram Dubey and Soumendu Chakraborty\\
\noindent\rule{\linewidth}{0.4pt}

\hfill \break

\noindent\textbf{Detailed Experimental Results on Different Trials:}\\
The tables in this supplementary material contain the experimental results for 30 trials under different training settings.

Table \ref{table:results_cifar} presents the experimental results for 30 trials over the original CIFAR10 training set using VGG16, ResNet56 and DenseNet121 models for different CIFAR10 test sets.

Table \ref{table:results_caltech} summarizes the classification accuracy for 30 trials over the original Caltech256 training set using VGG19, ResNet18 and DenseNet121 models for different Caltech256 test sets.

Table \ref{table:results_tinyimagenet} presents the experimental results on 30 trials over the original TinyImageNet training set using VGG16, ResNet101 and DenseNet121 models for different TinyImageNet test sets.

Table \ref{table:results_cifar-augment} presents the experimental results for 30 trials over the CCP augmented CIFAR10 training set using VGG16, ResNet56 and DenseNet121 models for different CIFAR10 test sets.

Table \ref{table:results_caltech-augment} summarizes the classification accuracy for 30 trials over the CCP augmented Caltech256 training set using VGG19, ResNet18 and DenseNet121 models for different Caltech256 test sets.

Table \ref{table:results_tinyimagenet-augment} presents the experimental results on 30 trials over the CCP augmented TinyImageNet training set using VGG16, ResNet101 and DenseNet121 models for different TinyImageNet test sets.

Table \ref{table:results_cifar-augment_others} presents the experimental results on 30 trials over the CCP augmented CIFAR10 training set using VGG16, ResNet56 and DenseNet121 models for One-pixel and Adversarial attacked CIFAR10 test sets.

Table \ref{table:results_caltech-augment_others} presents the experimental results on 30 trials over the CCP augmented Caltech256 training set using VGG19, ResNet18 and DenseNet121 models for One-pixel and Adversarial attacked Caltech256 test sets.

Table \ref{table:results_tinyimagenet-augment_others} presents the experimental results on 30 trials over the CCP augmented TinyImageNet training set using VGG16, ResNet101 and DenseNet121 models for One-pixel and Adversarial attacked TinyImageNet test sets.

Table \ref{table:results_comparison1_adv} presents the experimental results on 30 trials over the adversarial augmented training sets using different models for different test sets.

\begin{table}[!t]
\centering
\caption{Experimental results in terms of the accuracy ($\%$) over \textbf{original CIFAR10 training data} and different test sets. Thirty trials are performed using different CNN models, such as VGG16, ResNet56 and DenseNet121. The $CCP_f$ and $CCP_v$ represent the results using the proposed CCP attack with fixed and variable random weight settings, respectively. The STD denotes the standard deviation.}
\begin{tabular}{m{0.16\columnwidth}m{0.098\columnwidth}m{0.098\columnwidth}m{0.098\columnwidth}m{0.098\columnwidth}m{0.098\columnwidth}m{0.098\columnwidth}}
\toprule
\multirow{2}{*}{\textbf{Trials}} &\multicolumn{2}{c}{\textbf{VGG16}}
&\multicolumn{2}{c}{\textbf{ResNet56}} &\multicolumn{2}{c}{\textbf{DenseNet121}}\\
\cmidrule{2-7}
&$CCP_f$ &$CCP_v$ &$CCP_f$ &$CCP_v$ &$CCP_f$ &$CCP_v$\\\midrule
1 &64.34 &76.03 &79.99 &77.47 &57.44 &56.06 \\
2 &81.57 &75.81 &85.41 &77.65 &66.93 &56.59 \\
3 &76.11 &75.72 &81.76 &77.44 &49.31 &56.36 \\
4 &79.65 &75.86 &83.88 &77.73 &66.89 &56.89 \\
5 &78.07 &76.26 &66.50 &77.67 &57.95 &55.83 \\
6 &76.07 &76.12 &88.08 &77.49 &67.80 &56.58 \\
7 &70.56 &75.90 &82.75 &77.84 &72.65 &56.90 \\
8 &81.32 &76.30 &70.72 &77.57 &73.54 &56.09 \\
9 &81.69 &76.07 &81.65 &77.09 &40.35 &55.68 \\
10 &86.98 &75.82 &63.86 &77.11 &49.48 &55.83 \\
11 &85.61 &76.01 &81.99 &78.18 &60.30 &56.34 \\
12 &82.99 &75.94 &65.84 &77.86 &74.76 &56.78 \\
13 &80.53 &75.81 &81.35 &77.62 &61.96 &56.52 \\
14 &83.11 &76.06 &71.17 &76.94 &73.91 &56.71 \\
15 &78.18 &76.24 &79.95 &77.28 &73.15 &56.40 \\
16 &85.35 &75.98 &65.71 &77.68 &41.66 &56.34 \\
17 &67.95 &76.09 &59.99 &77.20 &54.97 &56.20 \\
18 &66.60 &75.69 &82.55 &77.44 &70.92 &56.62 \\
19 &77.54 &75.76 &83.40 &77.80 &65.61 &56.67 \\
20 &77.50 &75.96 &82.37 &77.27 &43.12 &56.11 \\
21 &86.76 &75.84 &74.41 &77.61 &65.05 &56.88 \\
22 &58.49 &76.24 &76.54 &77.67 &46.09 &56.99 \\
23 &53.59 &75.28 &83.01 &77.73 &37.77 &56.52 \\
24 &84.70 &75.48 &86.51 &77.76 &63.62 &55.92 \\
25 &71.94 &75.91 &83.95 &78.34 &61.00 &56.87 \\
26 &88.31 &75.39 &79.84 &77.62 &62.00 &56.78 \\
27 &83.03 &76.36 &80.48 &77.37 &57.48 &56.45 \\
28 &66.18 &76.25 &88.55 &77.64 &77.95 &56.49 \\
29 &81.85 &76.21 &75.42 &77.62 &57.01 &56.64 \\
30 &58.17 &75.20 &83.79 &77.72 &56.21 &56.64 \\
\hline
\textbf{Mean} &\textbf{76.49} &\textbf{75.92} &\textbf{78.38} &\textbf{77.58} &\textbf{60.23} &\textbf{56.46}\\
\textbf{STD} &9.16 &0.29 &7.54 &0.29 &10.93 &0.35\\
\textbf{Minimum} &53.59 &75.20 &59.99 &76.94 &37.77 &55.68\\
\textbf{Maximum} &88.31 &76.36 &88.55 &78.34 &77.95 &56.99\\
\hline
\textbf{Original Accuracy} &\multicolumn{2}{c}{\textbf{93.58}}
&\multicolumn{2}{c}{\textbf{91.44}} &\multicolumn{2}{c}{\textbf{92.72}}\\
\bottomrule
\end{tabular}
\label{table:results_cifar}
\end{table}

\newpage

\begin{table}[!htp]\centering
\caption{Experimental results in terms of the accuracy ($\%$) over \textbf{original Caltech256 training data} and different test sets. Thirty trials are performed using different CNN models, such as VGG19, ResNet18 and DenseNet121. The $CCP_f$ and $CCP_v$ represent the results using the proposed CCP attack with fixed and variable random weight settings, respectively. The STD denotes the standard deviation.}
\begin{tabular}{m{0.16\columnwidth}m{0.098\columnwidth}m{0.098\columnwidth}m{0.098\columnwidth}m{0.098\columnwidth}m{0.098\columnwidth}m{0.098\columnwidth}}
\toprule
\multirow{2}{*}{\textbf{Trials}} &\multicolumn{2}{c}{\textbf{VGG19}} &\multicolumn{2}{c}{\textbf{ResNet18}}
&\multicolumn{2}{c}{\textbf{DenseNet121}}\\
\cmidrule{2-7}
&$CCP_f$ &$CCP_v$ &$CCP_f$ &$CCP_v$ &$CCP_f$ &$CCP_v$\\\midrule
1 &16.14 &28.05 &38.94 &40.97 &38.88 &37.67\\
2 &30.65 &27.48 &27.84 &41.66 &39.47 &36.85\\
3 &31.66 &28.33 &45.00 &40.89 &49.32 &39.79\\
4 &29.36 &28.03 &41.24 &41.37 &44.60 &41.40\\
5 &30.71 &27.93 &22.84 &41.98 &36.35 &39.83\\
6 &22.18 &28.06 &44.91 &41.03 &39.54 &36.76\\
7 &16.57 &28.09 &47.12 &42.00 &30.01 &40.98\\
8 &35.49 &28.33 &47.46 &41.66 &30.99 &40.45\\
9 &33.45 &27.87 &29.48 &41.06 &42.19 &36.26\\
10 &33.64 &28.05 &51.70 &40.98 &36.03 &39.67\\
11 &21.73 &28.00 &33.17 &42.61 &46.79 &38.27\\
12 &40.55 &28.54 &29.86 &41.58 &29.99 &38.81\\
13 &25.91 &28.74 &40.28 &41.61 &42.03 &37.22\\
14 &33.94 &28.22 &41.74 &40.89 &30.15 &42.51\\
15 &20.23 &28.01 &37.45 &41.82 &37.16 &37.93\\
16 &16.26 &28.80 &22.31 &42.00 &47.44 &35.97\\
17 &38.64 &27.61 &46.37 &41.16 &26.73 &37.82\\
18 &33.14 &28.24 &42.49 &42.30 &40.04 &38.60\\
19 &31.37 &28.29 &29.62 &42.11 &42.45 &34.51\\
20 &36.10 &27.77 &28.78 &41.95 &40.79 &37.01\\
21 &26.47 &27.90 &14.77 &41.69 &27.77 &39.86\\
22 &23.43 &27.50 &43.70 &41.34 &38.78 &38.41\\
23 &26.31 &27.76 &49.44 &42.04 &42.96 &39.97\\
24 &28.29 &27.29 &50.63 &41.39 &39.76 &41.08\\
25 &26.23 &28.06 &38.93 &41.64 &34.14 &42.35\\
26 &24.24 &27.92 &22.39 &41.56 &32.50 &40.53\\
27 &38.14 &27.90 &37.67 &41.58 &45.50 &39.70\\
28 &35.05 &27.66 &44.12 &41.69 &31.37 &38.54\\
29 &23.82 &27.71 &40.12 &41.14 &41.72 &35.50\\
30 &33.12 &28.22 &49.95 &41.84 &37.46 &32.67\\\hline
\textbf{Mean} &\textbf{28.76} &\textbf{28.01} &\textbf{38.01} &\textbf{41.58} &\textbf{38.10} &\textbf{38.56}\\
\textbf{STD} &6.65 &0.36 &9.60 &0.43 &5.98 &2.25\\
\textbf{Minimum} &16.14 &27.29 &14.77 &40.89 &26.73 &32.67\\
\textbf{Maximum} &40.55 &28.80 &51.70 &42.61 &49.32 &42.51\\
\hline
\textbf{Original Accuracy} &\multicolumn{2}{c}{\textbf{50.42}} &\multicolumn{2}{c}{\textbf{66.73}} &\multicolumn{2}{c}{\textbf{64.83}}\\
\bottomrule
\end{tabular}
\label{table:results_caltech}
\end{table}

\newpage


\begin{table}[!htp]\centering
\caption{Experimental results in terms of the accuracy ($\%$) over \textbf{original TinyImagenet training data} and different test sets. Thirty trials are performed using different CNN models, such as VGG16, ResNet101 and DenseNet121. The $CCP_f$ and $CCP_v$ represent the results using the proposed CCP attack with fixed and variable random weight settings, respectively. The STD denotes the standard deviation.}
\begin{tabular}{m{0.16\columnwidth}m{0.098\columnwidth}m{0.098\columnwidth}m{0.098\columnwidth}m{0.098\columnwidth}m{0.098\columnwidth}m{0.098\columnwidth}}
\toprule
\multirow{2}{*}{\textbf{Trials}} &\multicolumn{2}{c}{\textbf{VGG16}}
&\multicolumn{2}{c}{\textbf{ResNet101}}
&\multicolumn{2}{c}{\textbf{DenseNet121}} \\
\cmidrule{2-7}
&$CCP_f$ &$CCP_v$ &$CCP_f$ &$CCP_v$ &$CCP_f$ &$CCP_v$\\\midrule
1 &15.48 &13.76 &27.23 &26.32 &30.10 &31.28 \\
2 &14.63 &13.93 &30.26 &27.08 &33.39 &31.41 \\
3 &9.34 &14.26 &21.18 &26.49 &26.10 &31.31 \\
4 &14.13 &14.39 &25.89 &26.65 &30.63 &30.78 \\
5 &20.12 &13.96 &33.48 &26.55 &36.01 &31.07 \\
6 &16.36 &13.86 &30.40 &26.21 &35.12 &31.20 \\
7 &13.71 &13.68 &26.59 &26.51 &30.72 &30.61 \\
8 &16.42 &13.96 &27.35 &27.18 &31.28 &31.09 \\
9 &11.07 &14.33 &24.35 &26.95 &34.34 &31.12 \\
10 &10.96 &14.64 &23.42 &27.05 &27.39 &31.14 \\
11 &3.34 &13.48 &19.07 &26.65 &26.86 &31.10 \\
12 &12.65 &14.19 &25.52 &26.71 &30.47 &30.59 \\
13 &13.91 &14.11 &25.82 &27.38 &30.05 &31.63 \\
14 &17.49 &14.11 &31.96 &26.86 &36.54 &31.41 \\
15 &13.58 &14.05 &26.74 &26.55 &29.90 &31.22 \\
16 &7.89 &14.38 &17.55 &26.82 &27.84 &30.78 \\
17 &12.06 &13.96 &25.28 &26.54 &28.73 &31.42 \\
18 &20.26 &14.00 &32.98 &26.36 &37.46 &30.89 \\
19 &13.70 &14.09 &25.60 &26.74 &29.45 &30.97 \\
20 &10.84 &13.85 &23.65 &26.41 &27.56 &31.09 \\
21 &15.30 &14.70 &28.66 &26.84 &31.80 &31.20 \\
22 &9.22 &14.49 &21.89 &26.57 &24.70 &31.12 \\
23 &13.57 &14.03 &25.29 &26.39 &30.21 &30.58 \\
24 &7.76 &14.35 &21.00 &26.58 &24.18 &31.10 \\
25 &12.20 &14.30 &23.84 &27.03 &28.06 &30.89 \\
26 &14.54 &13.84 &32.75 &26.94 &34.22 &31.06 \\
27 &18.70 &13.61 &31.10 &26.59 &35.79 &31.01 \\
28 &19.68 &13.91 &33.54 &27.06 &37.00 &31.25 \\
29 &14.87 &13.87 &27.25 &26.47 &30.70 &30.97 \\
30 &5.56 &13.90 &12.95 &27.15 &19.81 &31.45 \\
\hline
\textbf{Mean} &\textbf{13.31} &\textbf{14.07} &\textbf{26.09} &\textbf{26.72} &\textbf{30.55} &\textbf{31.09} \\
\textbf{STD} &4.04 &0.29 &4.81 &0.29 &4.09 &0.25 \\
\textbf{Minimum} &3.34 &13.48 &12.95 &26.21 &19.81 &30.58 \\
\textbf{Maximum} &20.26 &14.70 &33.54 &27.38 &37.46 &31.63 \\
\hline
\textbf{Original Accuracy} &\multicolumn{2}{c}{\textbf{42.55}}
&\multicolumn{2}{c}{\textbf{59.19}} &\multicolumn{2}{c}{\textbf{62.52}} \\
\bottomrule
\end{tabular}
\label{table:results_tinyimagenet}
\end{table}

\newpage

\begin{table}[!htp]\centering
\caption{Experimental results in terms of the accuracy ($\%$) over \textbf{CCP augmented CIFAR10 training data} and different test sets. Thirty trials are performed using different CNN models, such as VGG16, ResNet56 and DenseNet121.}
\begin{tabular}{m{0.16\columnwidth}m{0.098\columnwidth}m{0.098\columnwidth}m{0.098\columnwidth}m{0.098\columnwidth}m{0.098\columnwidth}m{0.098\columnwidth}}
\toprule
\multirow{2}{*}{\textbf{Trials}} &\multicolumn{2}{c}{\textbf{VGG16}}
&\multicolumn{2}{c}{\textbf{ResNet56}} &\multicolumn{2}{c}{\textbf{DenseNet121}}\\
\cmidrule{2-7}
&$CCP_f$ &$CCP_v$ &$CCP_f$ &$CCP_v$ &$CCP_f$ &$CCP_v$\\\midrule
1 &90.14 &90.81 &89.14 &89.34 &89.04 &88.54 \\
2 &91.18 &90.80 &88.69 &89.33 &88.30 &88.67 \\
3 &91.05 &90.43 &88.40 &89.52 &88.43 &88.81 \\
4 &91.26 &91.12 &89.00 &87.22 &89.46 &88.78 \\
5 &89.16 &91.30 &89.06 &89.38 &88.97 &88.66 \\
6 &90.69 &91.29 &88.69 &88.86 &88.49 &88.39 \\
7 &90.79 &91.16 &89.33 &88.94 &88.23 &88.88 \\
8 &91.04 &90.02 &89.25 &89.26 &89.17 &88.55 \\
9 &91.20 &91.19 &88.94 &89.17 &88.40 &88.66 \\
10 &91.18 &91.20 &88.34 &89.14 &88.72 &88.60 \\
11 &91.34 &91.42 &88.98 &89.44 &88.82 &88.70 \\
12 &89.52 &90.78 &89.20 &89.23 &89.19 &88.70 \\
13 &90.86 &91.08 &89.15 &89.40 &88.94 &88.95 \\
14 &90.87 &90.67 &89.26 &89.15 &89.20 &88.58 \\
15 &91.27 &91.38 &89.00 &86.63 &88.64 &88.67 \\
16 &91.35 &91.34 &89.14 &89.24 &88.75 &88.75 \\
17 &90.98 &88.09 &89.10 &88.91 &88.84 &88.59 \\
18 &91.13 &90.89 &89.60 &88.11 &89.48 &88.50 \\
19 &90.61 &91.06 &89.01 &89.37 &89.37 &88.72 \\
20 &90.97 &90.46 &88.59 &89.44 &89.00 &88.57 \\
21 &90.82 &91.43 &89.08 &89.23 &88.65 &88.60 \\
22 &90.86 &91.26 &89.16 &88.56 &89.28 &88.68 \\
23 &91.34 &90.86 &89.09 &88.44 &88.83 &88.52 \\
24 &91.30 &90.89 &89.50 &89.59 &88.49 &88.51 \\
25 &89.94 &91.28 &89.67 &88.67 &88.76 &88.67 \\
26 &90.97 &90.96 &89.41 &88.90 &88.68 &88.79 \\
27 &90.99 &90.87 &88.59 &88.33 &88.33 &88.69 \\
28 &91.01 &91.23 &88.43 &89.14 &89.18 &88.71 \\
29 &91.19 &90.60 &88.31 &89.43 &88.79 &88.86 \\
30 &89.24 &90.89 &88.95 &88.12 &88.34 &88.65 \\
\hline
\textbf{Mean} &\textbf{90.81} &\textbf{90.89} &\textbf{89.00} &\textbf{88.92} &\textbf{88.83} &\textbf{88.67} \\
\textbf{STD} &0.59 &0.61 &0.35 &0.67 &0.35 &0.12 \\
\textbf{Minimum} &89.16 &88.09 &88.31 &86.63 &88.23 &88.39 \\
\textbf{Maximum} &91.35 &91.43 &89.67 &89.59 &89.48 &88.95 \\
\hline
\textbf{Original Accuracy} &\multicolumn{2}{c}{\textbf{91.42}} &\multicolumn{2}{c}{\textbf{90.15}} &\multicolumn{2}{c}{\textbf{90.61}} \\
\bottomrule
\end{tabular}
\label{table:results_cifar-augment}
\end{table}

\newpage
\begin{table}[!htp]\centering
\caption{Experimental results in terms of the accuracy ($\%$) over \textbf{CCP augmented Caltech256 training data} and different test sets. Thirty trials are performed using different CNN models, such as VGG19, ResNet18 and DenseNet121.}
\begin{tabular}{m{0.16\columnwidth}m{0.098\columnwidth}m{0.098\columnwidth}m{0.098\columnwidth}m{0.098\columnwidth}m{0.098\columnwidth}m{0.098\columnwidth}}
\toprule
\multirow{2}{*}{\textbf{Trials}} &\multicolumn{2}{c}{\textbf{VGG19}} &\multicolumn{2}{c}{\textbf{ResNet18}}
&\multicolumn{2}{c}{\textbf{DenseNet121}}\\
\cmidrule{2-7}
&$CCP_f$ &$CCP_v$ &$CCP_f$ &$CCP_v$ &$CCP_f$ &$CCP_v$\\\midrule

1 &42.54 &40.02 &63.39 &63.55 &60.86 &60.48 \\
2 &38.73 &39.67 &63.11 &63.53 &60.37 &60.80 \\
3 &39.86 &40.12 &63.47 &63.61 &61.68 &60.75 \\
4 &41.71 &40.08 &63.66 &63.26 &62.41 &60.54 \\
5 &40.77 &40.04 &62.83 &63.36 &61.40 &60.58 \\
6 &34.39 &39.97 &63.71 &63.42 &62.18 &60.35 \\
7 &35.45 &40.15 &63.89 &63.08 &59.82 &60.45 \\
8 &42.61 &40.58 &61.30 &63.23 &57.28 &60.82 \\
9 &42.49 &39.95 &63.29 &63.53 &62.17 &60.95 \\
10 &39.20 &40.05 &63.36 &63.36 &57.41 &60.64 \\
11 &40.60 &40.36 &63.69 &62.86 &61.78 &60.51 \\
12 &31.00 &40.12 &63.03 &63.10 &60.66 &60.54 \\
13 &42.33 &40.12 &63.66 &63.42 &58.29 &60.67 \\
14 &37.93 &40.23 &63.84 &63.48 &60.96 &60.64 \\
15 &42.38 &39.63 &62.91 &63.11 &62.78 &60.74 \\
16 &36.03 &39.79 &60.72 &63.47 &58.68 &60.38 \\
17 &39.49 &39.70 &63.45 &63.00 &60.30 &60.62 \\
18 &44.15 &39.78 &64.03 &62.73 &62.70 &60.74 \\
19 &44.36 &40.28 &63.39 &63.60 &62.42 &60.77 \\
20 &35.12 &39.87 &63.34 &63.60 &62.62 &60.83 \\
21 &43.06 &39.95 &64.16 &63.32 &60.85 &60.50 \\
22 &42.40 &40.68 &63.08 &63.45 &61.75 &61.11 \\
23 &45.18 &40.07 &63.05 &63.53 &62.71 &60.64 \\
24 &44.58 &40.02 &63.66 &63.60 &62.65 &60.61 \\
25 &36.02 &40.41 &62.75 &62.95 &61.68 &60.48 \\
26 &36.10 &40.08 &63.81 &63.26 &59.58 &60.64 \\
27 &42.43 &39.70 &63.66 &63.65 &61.06 &60.56 \\
28 &44.52 &40.41 &62.65 &63.40 &60.88 &60.70 \\
29 &39.42 &40.26 &63.74 &63.56 &58.95 &60.46 \\
30 &39.41 &40.04 &63.87 &63.69 &60.96 &60.59 \\


\hline
\textbf{Mean} &\textbf{40.14} &\textbf{40.07} &\textbf{63.28} &\textbf{63.36} &\textbf{60.93} &\textbf{60.64} \\
\textbf{STD} &3.51 &0.25 &0.72 &0.24 &1.54 &0.16 \\
\textbf{Minimum} &31.00 &39.63 &60.72 &62.73 &57.28 &60.35 \\
\textbf{Maximum} &45.18 &40.68 &64.16 &63.69 &62.78 &61.11 \\
\hline
\textbf{Original Accuracy} &\multicolumn{2}{c}{\textbf{52.57}} &\multicolumn{2}{c}{\textbf{64.87}} &\multicolumn{2}{c}{\textbf{62.58}} \\
\bottomrule
\end{tabular}
\label{table:results_caltech-augment}
\end{table}

\newpage


\begin{table}[!htp]\centering
\caption{Experimental results in terms of the accuracy ($\%$) over \textbf{CCP augmented TinyImageNet training data} and different test sets. Thirty trials are performed using different CNN models, such as VGG16, ResNet101 and DenseNet121.}
\begin{tabular}{m{0.16\columnwidth}m{0.098\columnwidth}m{0.098\columnwidth}m{0.098\columnwidth}m{0.098\columnwidth}m{0.098\columnwidth}m{0.098\columnwidth}}
\toprule
\multirow{2}{*}{\textbf{Trials}} &\multicolumn{2}{c}{\textbf{VGG16}} &\multicolumn{2}{c}{\textbf{ResNet101}}
&\multicolumn{2}{c}{\textbf{DenseNet121}}\\
\cmidrule{2-7}
&$CCP_f$ &$CCP_v$ &$CCP_f$ &$CCP_v$ &$CCP_f$ &$CCP_v$\\\midrule
1 &35.02 &33.66 &53.37 &51.87 &56.49 &55.57 \\
2 &34.04 &33.68 &52.28 &51.61 &56.63 &55.37 \\
3 &32.43 &33.81 &49.19 &52.35 &53.33 &55.11 \\
4 &34.78 &33.80 &52.39 &51.65 &56.07 &55.43 \\
5 &35.79 &33.50 &52.30 &51.75 &55.91 &55.49 \\
6 &33.87 &33.89 &51.95 &51.88 &55.30 &55.72 \\
7 &34.08 &33.67 &52.70 &51.38 &56.58 &55.33 \\
8 &32.66 &33.88 &51.36 &52.15 &55.58 &55.58 \\
9 &35.05 &33.86 &52.40 &52.10 &56.92 &55.86 \\
10 &33.96 &33.95 &51.59 &51.59 &55.29 &55.25 \\
11 &33.15 &33.90 &51.53 &51.74 &55.91 &55.25 \\
12 &34.74 &33.80 &52.28 &51.52 &55.50 &55.41 \\
13 &32.26 &34.16 &50.87 &51.76 &55.18 &55.42 \\
14 &35.16 &33.78 &52.92 &51.63 &57.21 &55.40 \\
15 &35.03 &33.93 &52.33 &52.08 &55.66 &55.47 \\
16 &33.66 &34.00 &50.61 &51.88 &54.36 &55.81 \\
17 &33.93 &33.88 &53.21 &51.66 &56.37 &55.57 \\
18 &34.95 &34.47 &52.49 &52.04 &55.87 &55.73 \\
19 &34.27 &33.86 &53.28 &51.83 &57.10 &55.65 \\
20 &32.45 &33.86 &50.47 &51.84 &54.50 &55.55 \\
21 &33.76 &33.70 &51.86 &51.79 &55.78 &54.98 \\
22 &33.60 &34.31 &51.59 &51.53 &54.32 &55.18 \\
23 &34.95 &33.49 &53.37 &52.17 &56.99 &55.08 \\
24 &32.92 &33.98 &50.36 &51.94 &54.42 &55.34 \\
25 &32.57 &33.62 &50.30 &51.99 &54.14 &55.41 \\
26 &35.35 &33.79 &51.55 &52.16 &55.73 &55.69 \\
27 &34.66 &33.43 &53.15 &51.58 &56.71 &55.60 \\
28 &35.99 &33.56 &52.28 &52.20 &56.37 &55.82 \\
29 &33.56 &33.44 &51.64 &51.39 &55.31 &55.92 \\
30 &31.90 &33.97 &50.34 &52.09 &53.67 &55.61 \\
\hline
\textbf{Mean} &\textbf{34.02} &\textbf{33.82} &\textbf{51.87} &\textbf{51.84} &\textbf{55.64} &\textbf{55.49} \\
\textbf{STD} &1.09 &0.23 &1.05 &0.25 &1.02 &0.23 \\
\textbf{Minimum} &31.90 &33.43 &49.19 &51.38 &53.33 &54.98 \\
\textbf{Maximum} &35.99 &34.47 &53.37 &52.35 &57.21 &55.92 \\
\hline
\textbf{Original Accuracy} &\multicolumn{2}{c}{\textbf{43.73}} &\multicolumn{2}{c}{\textbf{59.65}} &\multicolumn{2}{c}{\textbf{63.41}} \\
\bottomrule
\end{tabular}
\label{table:results_tinyimagenet-augment}
\end{table}

\begin{table}[!htp]\centering
\caption{Experimental results in terms of the accuracy ($\%$) over \textbf{CCP augmented CIFAR10 training data} over different attacks. Thirty trials are performed using different CNN models, such as VGG16, ResNet56 and DenseNet121.}
\begin{tabular}{m{0.16\columnwidth}m{0.098\columnwidth}m{0.098\columnwidth}m{0.098\columnwidth}m{0.098\columnwidth}m{0.098\columnwidth}m{0.098\columnwidth}}
\toprule
\multirow{2}{*}{\textbf{Trials}} &\multicolumn{2}{c}{\textbf{VGG16}} &\multicolumn{2}{c}{\textbf{ResNet56}}
&\multicolumn{2}{c}{\textbf{DenseNet121}}\\
\cmidrule{2-7}
&$One Pixel$ &$Adversial$ &$One Pixel$ &$Adversial$  &$One Pixel$ &$Adversial$ \\\midrule
1 &90.47 &81.18 &88.99 &80.35 &89.01 &74.80 \\
2 &90.43 &81.36 &88.73 &81.45 &89.11 &74.91 \\
3 &90.67 &81.45 &89.34 &80.93 &89.17 &74.99 \\
4 &90.56 &81.45 &89.06 &80.99 &89.13 &75.38 \\
5 &90.62 &81.27 &88.94 &79.32 &89.07 &75.26 \\
6 &90.57 &81.27 &89.14 &80.44 &89.12 &75.01 \\
7 &90.58 &81.24 &89.24 &81.31 &89.09 &74.99 \\
8 &90.42 &81.37 &89.34 &80.56 &89.06 &75.09 \\
9 &90.50 &81.28 &88.73 &82.17 &89.14 &75.26 \\
10 &90.68 &81.55 &89.03 &79.88 &89.09 &75.26 \\
11 &90.60 &81.36 &88.94 &80.51 &89.22 &75.14 \\
12 &90.55 &81.64 &89.34 &80.17 &89.12 &75.04 \\
13 &90.56 &81.23 &89.14 &80.32 &89.21 &75.24 \\
14 &90.53 &81.55 &89.06 &80.68 &89.23 &75.40 \\
15 &90.57 &81.38 &89.14 &80.59 &88.98 &75.31 \\
16 &90.55 &81.14 &89.24 &79.23 &89.26 &74.95 \\
17 &90.68 &81.30 &88.73 &80.24 &88.95 &75.26 \\
18 &90.70 &81.18 &89.03 &81.05 &89.06 &74.98 \\
19 &90.57 &81.51 &88.94 &80.95 &88.97 &74.86 \\
20 &90.73 &81.24 &88.99 &80.82 &88.86 &75.11 \\
21 &90.59 &81.23 &89.14 &80.54 &89.06 &75.17 \\
22 &90.51 &81.26 &89.06 &80.05 &89.02 &75.10 \\
23 &90.45 &81.64 &89.24 &80.45 &89.07 &74.67 \\
24 &90.56 &81.14 &89.06 &81.40 &89.19 &75.22 \\
25 &90.46 &81.63 &89.34 &81.13 &89.37 &75.12 \\
26 &90.51 &80.92 &89.06 &80.54 &89.13 &74.58 \\
27 &90.57 &81.70 &88.94 &80.79 &89.38 &74.96 \\
28 &90.61 &81.41 &88.94 &80.02 &89.06 &75.55 \\
29 &90.63 &81.44 &89.25 &81.33 &89.25 &74.80 \\
30 &90.75 &81.39 &88.99 &80.48 &89.25 &74.98 \\
\hline

\textbf{Mean} &\textbf{90.57} &\textbf{81.36} &\textbf{89.07} &\textbf{80.62} &\textbf{89.12} &\textbf{75.08} \\
\textbf{STD} &0.08 &0.18 &0.17 &0.61 &0.12 &0.21 \\
\textbf{Minimum} &90.42 &80.92 &88.73 &79.23 &88.86 &74.58 \\
\textbf{Maximum} &90.75 &81.70 &89.34 &82.17 &89.38 &75.55 \\
\hline
\textbf{Original Accuracy} &\multicolumn{2}{c}{\textbf{91.41}} &\multicolumn{2}{c}{\textbf{90.15}} &\multicolumn{2}{c}{\textbf{90.61}} \\

\bottomrule
\end{tabular}
\label{table:results_cifar-augment_others}
\end{table}

\begin{table}[!htp]\centering
\caption{Experimental results in terms of the accuracy ($\%$) over \textbf{CCP augmented CALTECH256 training data} over different attacks. Thirty trials are performed using different CNN models, such as VGG19, ResNet18 and DenseNet121.}
\begin{tabular}{m{0.16\columnwidth}m{0.098\columnwidth}m{0.098\columnwidth}m{0.098\columnwidth}m{0.098\columnwidth}m{0.098\columnwidth}m{0.098\columnwidth}}
\toprule
\multirow{2}{*}{\textbf{Trials}} &\multicolumn{2}{c}{\textbf{VGG19}} &\multicolumn{2}{c}{\textbf{ResNet18}}
&\multicolumn{2}{c}{\textbf{DenseNet121}}\\
\cmidrule{2-7}
&$One Pixel$ &$Adversial$ &$One Pixel$ &$Adversial$  &$One Pixel$ &$Adversial$ \\\midrule
1 &52.33 &49.47 &66.67 &64.66 &62.44 &60.29 \\
2 &52.30 &49.53 &66.57 &64.69 &62.50 &60.22 \\
3 &52.38 &49.39 &66.60 &64.83 &62.49 &60.64 \\
4 &52.33 &49.47 &66.62 &64.35 &62.50 &60.32 \\
5 &52.31 &49.66 &66.70 &64.66 &62.49 &60.30 \\
6 &52.43 &49.53 &66.62 &64.64 &62.46 &60.37 \\
7 &52.30 &49.53 &66.70 &64.80 &62.46 &60.37 \\
8 &52.33 &49.45 &66.60 &64.72 &62.44 &60.06 \\
9 &52.31 &49.60 &66.60 &64.72 &62.58 &60.29 \\
10 &52.33 &49.60 &66.57 &64.74 &62.47 &60.53 \\
11 &52.30 &49.58 &66.68 &64.77 &62.41 &60.54 \\
12 &52.35 &49.45 &66.60 &64.85 &62.52 &60.29 \\
13 &52.36 &49.57 &66.59 &64.43 &62.47 &60.40 \\
14 &52.35 &49.57 &66.63 &64.61 &62.52 &60.17 \\
15 &52.33 &49.37 &66.67 &64.63 &62.52 &60.50 \\
16 &52.31 &49.37 &66.67 &64.72 &62.49 &60.25 \\
17 &52.31 &49.53 &66.63 &64.88 &62.46 &60.17 \\
18 &52.30 &49.57 &66.65 &64.63 &62.55 &60.35 \\
19 &52.27 &49.86 &66.63 &64.82 &62.42 &60.32 \\
20 &52.31 &49.58 &66.76 &64.90 &62.57 &60.40 \\
21 &52.33 &49.52 &66.71 &64.42 &62.47 &60.22 \\
22 &52.31 &49.63 &66.73 &64.79 &62.49 &60.32 \\
23 &52.36 &49.61 &66.63 &64.61 &62.49 &60.38 \\
24 &52.36 &49.29 &66.60 &64.82 &62.50 &60.38 \\
25 &52.33 &49.53 &66.59 &64.61 &62.52 &60.62 \\
26 &52.35 &49.71 &66.67 &64.64 &62.49 &60.62 \\
27 &52.30 &49.61 &66.63 &64.72 &62.42 &60.24 \\
28 &52.38 &49.66 &66.59 &64.87 &62.52 &60.45 \\
29 &52.33 &49.58 &66.75 &64.61 &62.55 &60.27 \\
30 &52.33 &49.57 &66.59 &64.67 &62.44 &60.40 \\
\hline

\textbf{Mean} &\textbf{52.33} &\textbf{49.55} &\textbf{66.64} &\textbf{64.69} &\textbf{62.49} &\textbf{60.36} \\
\textbf{STD} &0.03 &0.11 &0.05 &0.13 &0.04 &0.14 \\
\textbf{Minimum} &52.27 &49.29 &66.57 &64.35 &62.41 &60.06 \\
\textbf{Maximum} &52.43 &49.86 &66.76 &64.90 &62.58 &60.64 \\
\hline
\textbf{Original Accuracy} &\multicolumn{2}{c}{\textbf{52.57}} &\multicolumn{2}{c}{\textbf{66.73}} &\multicolumn{2}{c}{\textbf{62.58}} \\
\bottomrule
\end{tabular}
\label{table:results_caltech-augment_others}
\end{table}

\begin{table}[!htp]\centering
\caption{Experimental results in terms of the accuracy ($\%$) over \textbf{CCP augmented TinyImagenet training data} over different attacks. Thirty trials are performed using different CNN models, such as VGG16, ResNet101 and DenseNet121.}
\begin{tabular}{m{0.16\columnwidth}m{0.098\columnwidth}m{0.098\columnwidth}m{0.098\columnwidth}m{0.098\columnwidth}m{0.098\columnwidth}m{0.098\columnwidth}}
\toprule
\multirow{2}{*}{\textbf{Trials}} &\multicolumn{2}{c}{\textbf{VGG16}} &\multicolumn{2}{c}{\textbf{ResNet101}}
&\multicolumn{2}{c}{\textbf{DenseNet121}}\\
\cmidrule{2-7}
&$One Pixel$ &$Adversial$ &$One Pixel$ &$Adversial$  &$One Pixel$ &$Adversial$ \\\midrule
1 &43.63 &41.01 &59.72 &56.83 &63.05 &60.63 \\
2 &43.61 &40.70 &59.62 &56.89 &63.12 &60.67 \\
3 &43.54 &40.82 &59.47 &57.02 &63.25 &60.60 \\
4 &43.66 &40.50 &59.61 &57.12 &63.11 &60.59 \\
5 &43.40 &40.81 &59.46 &56.92 &63.00 &60.37 \\
6 &43.67 &41.07 &59.47 &57.02 &63.22 &60.56 \\
7 &43.58 &40.91 &59.63 &56.95 &62.96 &60.53 \\
8 &43.57 &40.87 &59.43 &57.15 &63.11 &60.29 \\
9 &43.57 &41.02 &59.62 &56.68 &62.97 &60.46 \\
10 &43.56 &40.80 &59.67 &56.69 &63.24 &60.56 \\
11 &43.63 &41.10 &59.43 &57.00 &63.12 &60.89 \\
12 &43.72 &40.95 &59.55 &56.95 &63.09 &60.71 \\
13 &43.49 &40.82 &59.69 &56.81 &63.09 &60.19 \\
14 &43.45 &40.80 &59.57 &56.73 &63.26 &60.42 \\
15 &43.56 &40.93 &59.70 &56.70 &63.25 &60.61 \\
16 &43.58 &40.94 &59.62 &56.79 &63.03 &60.57 \\
17 &43.56 &40.88 &59.39 &56.72 &63.10 &60.72 \\
18 &43.54 &41.00 &59.66 &56.86 &63.13 &60.07 \\
19 &43.54 &40.95 &59.71 &56.94 &63.20 &60.19 \\
20 &43.38 &40.86 &59.53 &57.09 &63.28 &60.61 \\
21 &43.62 &40.92 &59.58 &57.01 &63.15 &60.46 \\
22 &43.35 &40.94 &59.62 &56.77 &63.17 &60.71 \\
23 &43.66 &40.54 &59.60 &56.33 &63.01 &60.33 \\
24 &43.55 &40.67 &59.62 &56.78 &63.27 &60.68 \\
25 &43.44 &40.97 &59.54 &57.03 &63.41 &60.60 \\
26 &43.52 &40.97 &59.63 &56.77 &63.23 &60.80 \\
27 &43.54 &40.74 &59.63 &57.25 &63.16 &60.53 \\
28 &43.72 &40.68 &59.48 &57.02 &63.03 &60.60 \\
29 &43.57 &40.98 &59.66 &56.60 &63.02 &60.42 \\
30 &43.51 &40.80 &59.72 &57.09 &63.21 &60.58 \\
\hline

\textbf{Mean} &\textbf{43.55} &\textbf{40.87} &\textbf{59.59} &\textbf{56.88} &\textbf{63.14} &\textbf{60.53} \\
\textbf{STD} &0.09 &0.14 &0.09 &0.19 &0.11 &0.18 \\
\textbf{Minimum} &43.35 &40.50 &59.39 &56.33 &62.96 &60.07 \\
\textbf{Maximum} &43.72 &41.10 &59.72 &57.25 &63.41 &60.89 \\
\hline
\textbf{Original Accuracy} &\multicolumn{2}{c}{\textbf{43.73}} &\multicolumn{2}{c}{\textbf{59.65}} &\multicolumn{2}{c}{\textbf{63.41}} \\

\bottomrule
\end{tabular}
\label{table:results_tinyimagenet-augment_others}
\end{table}

\begin{table*}[!t]
\caption{The comparison of results for different attacks on the original CIFAR10, Caltech256 and TinyImageNet test sets using VGG, ResNet and DenseNet Models adversial augmentation in training data. The results of the proposed CCP attack are reported under both fixed and variable random weight settings. These results are computed as an average and standard deviation over $30$ trials. The \% improvement in accuracy is also mentioned after CCP augmentation of training data w.r.t. the accuracy without training augmentation as presented in Table IV in the main paper.
The $\uparrow$ and $\downarrow$ represent the gain and loss, respectively.}
\centering
\resizebox{\columnwidth}{!}{%
\begin{tabular}{m{3.2cm}m{3.6cm}m{3.6cm}m{3.6cm}}
\hline
\multicolumn{4}{c}{\textbf{Accuracy over CIFAR10 Dataset}}\\
\hline
\textbf{Type of Attack} & \textbf{VGG16} & \textbf{ResNet56} & \textbf{DenseNet121}  \tabularnewline
\hline
Without Attack & $92.99$ $\pm$  $0.00$ ($\downarrow$ $0.63\%$) & $92.13$ $\pm$  $0.00$ ($\uparrow$ $0.75\%$) & $91.99$ $\pm$  $0.00$ ($\downarrow$ $0.78\%$)\tabularnewline
One Pixel Attack& $92.23$ $\pm$  $0.09$ ($\downarrow$ $0.62\%$)& $91.32$  $\pm$ $0.10$ ($\uparrow$ $2.14\%$)& $91.29$ $\pm$  $0.08$ ($\downarrow$ $0.99\%$)\tabularnewline
Adversarial Attack & $91.51$ $\pm$  $0.14$ ($\uparrow$ $10.77\%$)& $90.94$ $\pm$ $0.13$ ($\uparrow$ $13.98\%$)& $89.97$ $\pm$ $0.14$ ($\uparrow$ $33.42\%$)\tabularnewline
CCP Attack ($CCP_f$) & $76.02$ $\pm$ $8.23$ ($\downarrow$ $0.61\%$)& $64.21$ $\pm$ $8.27$ ($\downarrow$ $18.07\%$)&$61.28$ $\pm$ $11.7$ ($\uparrow$ $1.74\%$)\tabularnewline
CCP Attack ($CCP_v$) & $73.37$ $\pm$ $0.36$ ($\downarrow$ $3.35\%$)& $64.26$ $\pm$ $0.32$ ($\downarrow$ $17.16\%$)& $57.80$ $\pm$ $0.43$ ($\uparrow$ $2.37\%$)\tabularnewline
\hline
\multicolumn{4}{c}{\textbf{Accuracy over Caltech256 Dataset}}\\
\hline
\textbf{Type of Attack} & \textbf{VGG19} & \textbf{ResNet18} & \textbf{DenseNet121}  \tabularnewline
\hline
Without Attack & $53.10$ $\pm$ $0.00$ ($\uparrow$ $5.31\%$)& $65.43$ $\pm$ $0.00$ ($\downarrow$ $1.94\%$)& $66.41$ $\pm$ $0.00$ ($\uparrow$ $2.43\%$)\tabularnewline
One Pixel Attack& $53.15$ $\pm$  $0.05$  ($\uparrow$ $5.39\%$)& $65.33$  $\pm$ $0.05$ ($\downarrow$ $1.96\%$)& $66.43$ $\pm$  $0.04$ ($\uparrow$ $2.53\%$)\tabularnewline
Adversarial Attack & $52.09$ $\pm$  $0.13$ ($\uparrow$ $8.27\%$)& $65.03$ $\pm$ $0.15$ ($\uparrow$ $0.52\%$)& $65.23$ $\pm$  $0.14$ ($\uparrow$ $2.57\%$)\tabularnewline
CCP Attack ($CCP_f$) & $31.17$ $\pm$  $6.85$ ($\uparrow$ $8.37\%$)& $42.54$  $\pm$ $8.07$ ($\uparrow$ $11.91\%$)& $51.95$ $\pm$  $3.47$ ($\uparrow$ $36.35\%$)\tabularnewline
CCP Attack ($CCP_v$) & $31.12$ $\pm$  $0.34$ ($\uparrow$ $11.10\%$)& $42.88$ $\pm$ $0.37$ ($\uparrow$ $3.12\%$)& $52.18$ $\pm$  $0.32$ ($\uparrow$ $35.32\%$)\tabularnewline
\hline
\multicolumn{4}{c}{\textbf{Accuracy over TinyImageNet Dataset}}\\
\hline
\textbf{Type of Attack} & \textbf{VGG16} & \textbf{ResNet101} & \textbf{DenseNet121}  \tabularnewline
\hline
Without Attack & $43.62$ $\pm$ $0.00$ ($\uparrow$ $2.51\%$)& $59.30$ $\pm$ $0.00$ ($\uparrow$ $0.18\%$)& $62.27$ $\pm$  $0.00$ ($\downarrow$ $0.40\%$)\tabularnewline
One Pixel Attack & $43.31$ $\pm$ $0.08$ ($\uparrow$ $1.64\%$)& $59.25$ $\pm$ $0.10$ ($\uparrow$ $0.40\%$)& $62.09$ $\pm$  $0.08$ ($\downarrow$ $0.56\%$)\tabularnewline
Adversarial Attack & $43.43$ $\pm$ $0.17$ ($\uparrow$ $7.87\%$)& $59.32$ $\pm$ $0.16$ ($\uparrow$ $5.10\%$)& $62.26$ $\pm$  $0.17$ ($\uparrow$ $3.74\%$)\tabularnewline
CCP Attack ($CCP_f$) & $13.25$ $\pm$ $4.52$ ($\downarrow$ $0.45\%$)& $28.53$ $\pm$ $4.45$ ($\uparrow$ $9.35\%$)& $30.84$ $\pm$  $4.70$ ($\uparrow$ $0.94\%$) \tabularnewline
CCP Attack ($CCP_v$) & $14.13$ $\pm$ $0.21$ ($\uparrow$ $0.42\%$)& $29.36$ $\pm$ $0.30$ ($\uparrow$ $9.88\%$)& $31.50$ $\pm$  $0.21$ ($\uparrow$ $1.31\%$)\tabularnewline
\hline
\end{tabular}
}
\label{table:results_comparison1_adv}
\end{table*}

\end{document}